%% file: main.tex
\documentclass{article}

\PassOptionsToPackage{numbers, compress}{natbib}

 \usepackage[preprint]{neurips_2026}

\input{preamble}

\title{\boldcoolname: Spherical Priors for Plug-and-Play Restoration}

\author{%
  Sean Man \\
  Technion -- Israel Institute of Technology \\
  \texttt{sean.man@campus.technion.ac.il} \\
  \And
  Ron Raphaeli\thanks{denotes equal contribution} \\
  Independent Researcher \\
  \hphantom{\texttt{sean.man@campus.technion.ac.il}} \\
  \AND
  Matan Kleiner\footnotemark[1] \\
  Technion -- Israel Institute of Technology \\
  \texttt{matankleiner@campus.technion.ac.il} \\
  \And
  Or Ronai \\
  Technion -- Israel Institute of Technology \\
  \texttt{or.ronai@campus.technion.ac.il} \\
}

\begin{document}

\maketitle

\input{sections/0_abstract}

\input{sections/1_intro}

\input{sections/2_background}

\input{sections/4_method}

\input{sections/5_experiments}

\input{sections/6_conclusion}

\clearpage

\begin{ack}
The authors acknowledge Modal Labs for providing some of the cloud computing platform and GPUs resources ($8\times$ NVIDIA L40S) used for conducting part of the experiments presented in this paper.
\end{ack}

\bibliographystyle{unsrt}
\bibliography{citations}

\clearpage

\renewcommand\thefigure{S\arabic{figure}}    
\setcounter{figure}{0}  
\renewcommand{\thesection}{\Alph{section}}
\setcounter{section}{0}
\renewcommand{\thetable}{S\arabic{table}}
\setcounter{table}{0}
\renewcommand{\theequation}{S\arabic{equation}}
\setcounter{equation}{0}

\appendix

\crefalias{section}{appendix}

\input{sections/B_comparisons}

\clearpage
\input{sections/C_tables}

\clearpage
\input{sections/D_implementation}

\clearpage
\input{sections/A_proofs_v2}

\end{document}

%% file: preamble.tex
\usepackage[utf8]{inputenc} %
\usepackage[T1]{fontenc}    %
\usepackage{hyperref}       %
\usepackage{url}            %
\usepackage{booktabs}       %
\usepackage{amsfonts}       %
\usepackage{amssymb}
\usepackage{bm}
\usepackage{nicefrac}       %
\usepackage{microtype}      %
\usepackage[table,xcdraw, dvipsnames]{xcolor}
\usepackage{amsmath}
\usepackage[ruled,vlined,linesnumbered]{algorithm2e}
\usepackage{graphicx}
\usepackage{subcaption}
\usepackage[capitalize,noabbrev]{cleveref}
\usepackage{wrapfig}
\usepackage{tikz}

\usepackage[ruled,vlined]{algorithm2e} 
\usepackage{caption} %

\hypersetup{
	hidelinks,
	colorlinks=true,
	linkcolor=Blue,
	filecolor=Blue,
	citecolor=Blue,
	urlcolor=Blue,
	breaklinks=false
}

\SetKwInput{KwInput}{Input}
\SetKwInput{KwOutput}{Output}

\newcommand{\coolname}{SP$^3$\xspace}
\newcommand{\boldcoolname}{\textbf{SP}$\bm{^3}$\xspace}

\crefname{appendix}{Appendix}{Appendices}
\Crefname{appendix}{Appendix}{Appendices}

%% file: sections/0_abstract.tex
\begin{abstract}
  In this paper, we introduce \coolname, a novel Plug-and-Play algorithm that accelerates maximum a posteriori image restoration by replacing denoisers with Spherical Encoders (SE) as generative priors. \coolname approximates the intractable proximal prior step by utilizing the SE tightly structured latent space as a robust projection onto the natural image manifold. 
  Alternating this projection with a closed-form data-consistency step, via Half-Quadratic Splitting, achieves stable convergence without requiring gradient computation during inference. 
  This unique formulation unlocks ``anytime'' restoration capabilities, producing sharp, plausible images from the first iteration. Evaluations across a variety of image restoration tasks demonstrate that \coolname achieves perceptual quality comparable to state-of-the-art zero-shot diffusion and flow methods while being $3$-$630\times$ faster.
\end{abstract}

%% file: sections/1_intro.tex
\section{Introduction}
\label{sec:intro}

Recovering clean images from degraded measurements is a fundamental task in image processing, as photos are often acquired as degraded.
A popular technique for solving image restoration is by optimizing the maximum a posteriori (MAP) objective~\cite{PlugPlayPriorsModel2013venkatakrishnan}. MAP estimator maximizes the log-likelihood of a data fidelity term and the log-probability of a natural image prior term.  
After maximizing these two objectives, the resulted image is both natural looking and loyal to the measurements.

\begin{figure}[h]
  \centering
  \begin{minipage}[t]{0.49\linewidth}
    \vspace{0pt}
    \centering
    \includegraphics[width=\linewidth]{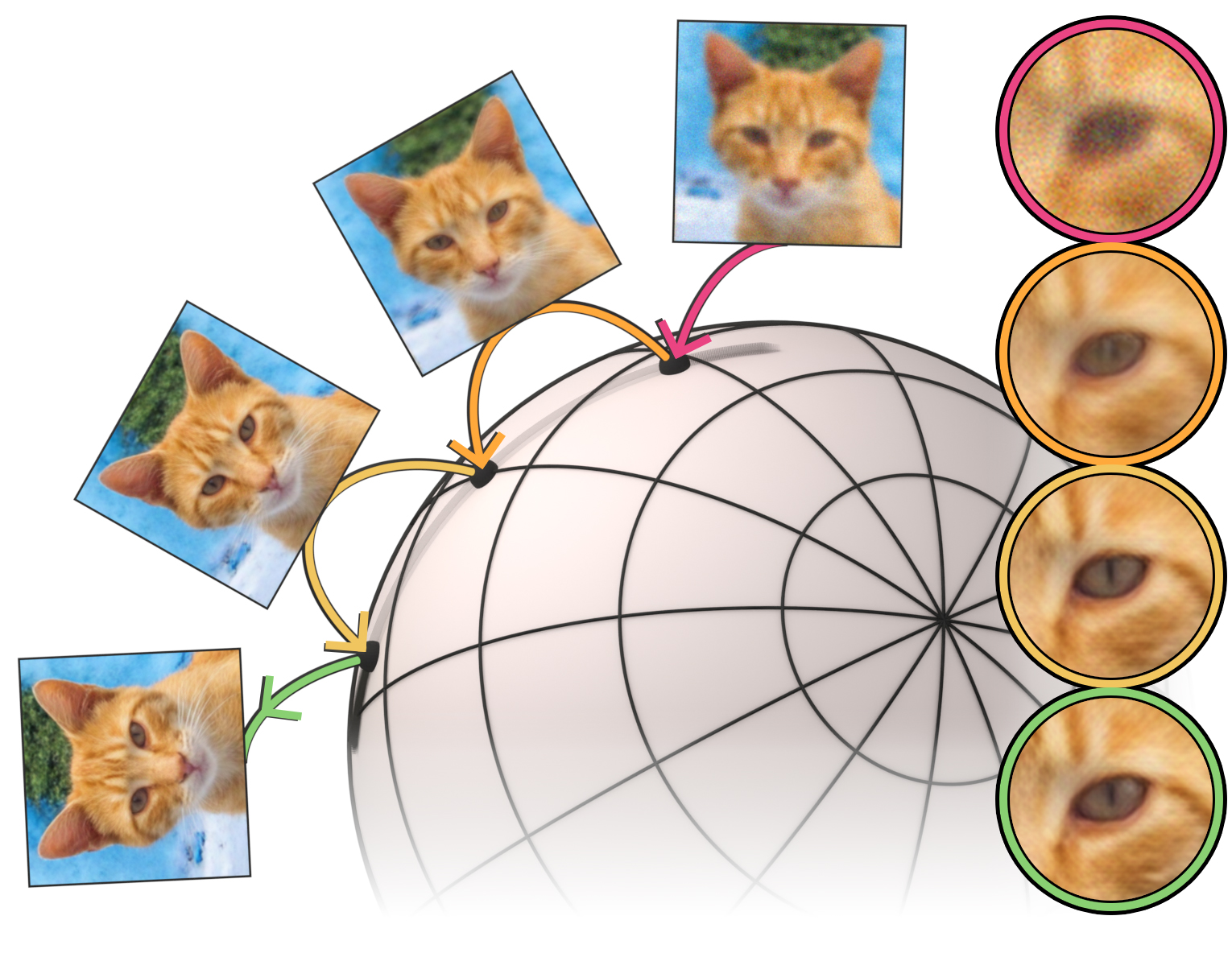}
  \end{minipage}
  \hfill
  \begin{minipage}[t]{0.49\linewidth}
    \vspace{0pt}
    \centering
    \includegraphics[width=\linewidth]{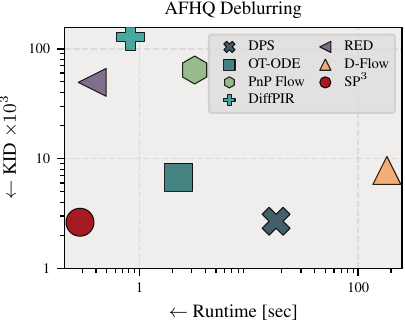}
  \end{minipage}
  \caption{
  \boldcoolname. We introduce a Plug-and-Play image restoration algorithm \emph{(left)} that iteratively projects degraded observations onto a tightly structured latent space using a Spherical Encoder (SE) generative prior. The trajectory along the sphere illustrates our gradient-free, alternating optimization approach, which navigates the image manifold to enforce data consistency. The magnified crops highlight the algorithm's ``anytime'' generation capability, demonstrating how it stably converges to produce perceptually sharp and plausible details from the very first iteration.
  \emph{(right)} \coolname achieves perceptual quality comparable to state-of-the-art zero-shot methods, while being $3-630\times$ faster.
  }
\label{fig:hero}
\end{figure}

Although the data-fidelity term is often straightforward to obtain, specifying an explicit, tractable prior function is challenging. As a result, many approaches have been developed to replace the prior term with easy-to-compute alternatives~\cite{PlugPlayPriorsModel2013venkatakrishnan, LittleEngineThat2017romanoa}. One prominent example is the Plug-and-Play (PnP) restoration framework~\cite{PlugPlayPriorsModel2013venkatakrishnan}, which replaces the explicit prior optimization step within an iterative algorithm with a denoising operator. 
This framework is particularly attractive because it allows using arbitrary pretrained denoisers. Consequently, better denoisers, such as neural and generative denoisers, can be incorporated with minimal modifications to the underlying reconstruction algorithm~\cite{zhang2017learning, PlugPlayImageRestoration2022zhang, DenoisingDiffusionModels2023zhua, martin2025pnpflow}.

With improvements in generative modeling, early methods with a structured latent space, such as generative adversarial network (GAN)~\cite{goodfellow_generative_2014}, variational autoencoders (VAE)~\cite{kingma2013auto}, and normalizing flows~\cite{dinh2017density}, were used as an image prior in restoration tasks~\cite{bora2017compressed, hand2018phase, asim2020invertible, menon2020pulse, whang2021solving, wei2022deep, prost2023inverse}. 
These methods often solve MAP optimization problem in the models' latent space, where the prior term is known and easy to sample from, such as a Gaussian. 
However, they mostly yield poor results~\cite{martin2025pnpflow}.  
Recent work focuses on utilizing diffusion and flow models~\cite{DeepUnsupervisedLearning2015sohl-dickstein, DenoisingDiffusionProbabilistic2020hoa, DenoisingDiffusionImplicit2020song, song2021scorebased, liu2023flow, lipman2023flow, albergo2025stochastic} as generative priors for zero-shot image restoration~\cite{DenoisingDiffusionRestoration2022kawara, DiffusionPosteriorSampling2022chunga, ZeroShotImageRestoration2022wangb, PseudoinverseGuidedDiffusionModels2022songa, SolvingLinearInverserout2023, PGDiffGuidingDiffusion2023yang, DifFaceBlindFace2023yue,  pokle2024trainingfree, ben2024d, zhang2024flow, Raphaeli_2025_ICCV, man2025elad, garber2025zero}. Yet, their reliance on gradients computed during inference, combined with their iterative nature, significantly increases restoration time.

In this work, we use the recently proposed Sphere Encoder (SE)~\cite{yue2026image}, a generative model that maps images onto a latent spherical manifold (and vice versa) as a prior for image restoration. 
Following the original motivation behind Plug-and-Play, 
enforcing a prior information through a projection operator, 
we design \textbf{Spherical Prior for Plug-and-Play Restoration (\coolname)}. \coolname, schematically illustrated in the left part of \cref{fig:hero}, alternates, via Half-Quadratic Splitting (HQS), between a prior step and a data fidelity step. The prior step projects the current estimate onto the learned spherical manifold, refines the resulting latent representation, and projects it back to the image manifold, where the data-fidelity step applies a data proximal operator.
Importantly, our method stably converges without requiring gradient computation or backpropagation at all. 

Unlike diffusion/flow-based methods, \coolname is extremely fast, achieving $\times 3-630$ speed improvement compared to previous work, illustrated in the right part of \cref{fig:hero}.
Moreover, one attractive property of \coolname, emerges from using the SE prior, is anytime restoration -- producing a single step reconstruction that could be further refined to improve its perceptual quality (left part of \cref{fig:hero}).
\coolname achieves these fast and anytime results while producing results on par to different PnP and diffusion/flow based methods, across different noisy restoration problems.

%% file: sections/2_background.tex
\section{Background and related work}
\label{sec:background}

\subsection{Sphere Encoder} 

Sphere Encoder (SE) \cite{yue2026image} is a recent generative framework that, similar to VAEs~\cite{kingma2013auto}, relies on an autoencoder~\cite{hinton2006reducing} to encode images onto a structured latent space and decode latent vectors back to the image space. 
Unlike VAEs that promote the latent space to follow a Gaussian distribution, SE promotes the latent space to be uniformly distributed on a sphere. 
By inducing a latent sphere, SE overcomes many of the shortcomings associated with VAEs, especially the posterior hole problem~\cite{dai2018diagnosing,rezende2018taming,makhzani2016advae}, which results in unrealistic image samples. 

To create the spherical latent space, SE employs a spherifying function (RMS normalization), denoted by $f(\cdot)$, to the output of the encoder $E$. Namely, a latent vector is given by $v=f(E(x))$. 
During training, instead of decoding the latent vector $v$ with the decoder $D$, the SE framework first add noise to the latent vector, followed by spherification, $S(v,\sigma)=f(v+\sigma\cdot e)$, where 
$e \sim \mathcal{N}(0,I)$ is a white Gaussian noise vector and $\sigma$ is a scalar that controls the noise magnitude. 
Adding noise to the latent vectors during training results in a latent sphere that is entirely covered and densely populated.

The training objective of SE compose of three reconstruction loss functions:
\begin{equation}\label{eq:se}
    \mathcal{L}_{\text{SE}}=\mathcal{L}_{\text{L1+perc.}} (D(v_{n}), x)+\mathcal{L}_{\text{L1+perc.}} (D(v_{N}), \text{sg}(D(v_{n}))) + \mathcal{L}_{\text{cos.~sim.}} (v, E(D(v_{N}))),
\end{equation}
where L1 denotes smoothed L1 loss~\cite{girshick2015fast}, perc.~denotes perceptual loss~\cite{gatys2016image}, cos.~sim~denotes cosine similarity, $v_\epsilon=v+\epsilon$, and $n=\sigma_1 e$, $N=\sigma_2 e$ with $\sigma_2 > \sigma_1$.  

In this work, we use the SE framework as a prior for image restoration, and illustrate how its unique features are crucial to the success of \coolname.

\subsection{Zero-shot generative priors for image restoration}

\begin{wrapfigure}{R}{0.5\textwidth}
  \centering
  \vspace{-1.3em} 
  \captionof{table}{\textbf{SE baseline quantitative comparisons.}
  Comparisons between \coolname~and baselines that solve MAP optimization problem in SE latent space. 
  Results are given for deblurring on AFHQ-Cat. We highlight the \colorbox[HTML]{FFCCC9}{best} method in each metric.}
  \begin{tabular}{lccc}
    \hline
    & S-GD & S-GPD & \coolname \\
    \hline
    PSNR {[}dB{]} $\uparrow$ & 18.5 & 23.1 & \colorbox[HTML]{FFCCC9}{25.7} \\
    LPIPS $\downarrow$ & 0.58 & 0.26 & \colorbox[HTML]{FFCCC9}{0.12} \\
    KID$\times10^3$ $\downarrow$ & 34.3 & 5.49 & \colorbox[HTML]{FFCCC9}{2.64} \\
    Time {[}sec{]} & 1.32 & 1.17 & \colorbox[HTML]{FFCCC9}{0.28} \\
    \hline
  \end{tabular}
  \label{tab:baselines}
  \vspace{0.5em} 
\end{wrapfigure}

Image restoration aims to recover a clean natural image, $\hat{x}$, from a degraded measurement $y$. 
The degraded measurement is considered to be the result of a degradation operator, $A$, applied to an unknown clean image $x$, with the addition of white Gaussian noise $n\sim\mathcal{N}(0,\sigma^2_n I)$, 
\begin{equation}\label{eq:degrdation}
    y = Ax + n.
\end{equation}
The degradation operator $A$ is linear for many common degradation types, such as inpainting, super-resolution, deblurring, and denoising.  
Following the notations of Blau \& Michaeli~\cite{PerceptionDistortionTradeoff2018blaua}, the natural image $x$ is a realization of a random variable $X$ with probability density function $p_X$. The measurement $y$ is a realization of a random variable $Y$, which relates to $X$ via the conditional probability density function $\smash{p_{Y|X}}$.
In general, an image restoration algorithm is an estimator $\hat{X}$ that generates reconstructions according to $\smash{p_{\hat{X}|Y}}$, where $X$ and $\smash{\hat{X}}$ are statistically independent given $Y$. 

The success of deep learning in image processing tasks led to a wide adoption of deep networks for image restoration. 
Early approaches trained regression based networks for image restoration~\cite{dong2015image, sun2015learning, zhang2016colorful, GaussianDenoiserResidual2017zhang, kligvasser2018xunit, ResidualDenseUNet2021gurrola-ramos, SwinIRImageRestoration2021liangb}, while recent approaches trained conditional generative models~\cite{ledig2017photo, SRFlowLearningSuperResolution2020lugmayr, RealESRGANTrainingRealWorld2021wangb, HighPerceptualQuality2021ohayona, whang_deblurring_2022, HighPerceptualQualityJPEG2023man, saharia_palette_2022, VQFRBlindFace2022gu, RobustBlindFace2022zhou, lugmayr2022repaint, delbracio2023inversion, DiffBIRBlindImage2024lin, PosteriorMeanRectifiedFlow2024ohayon, elata2025invfusion}. 
Despite the impressive results of these methods, they are often constrained by degradation processes seen during training.

An alternative, flexible, approach leverages pretrained models as image priors for zero-shot image restoration. Frameworks such as PnP~\cite{PlugPlayPriorsModel2013venkatakrishnan} and RED~\cite{LittleEngineThat2017romanoa} demonstrate that off-the-shelf deep~\cite{zhang2017learning, PlugPlayImageRestoration2022zhang} and generative~\cite{DenoisingDiffusionModels2023zhua, VariationalPerspectiveSolving2023mardani, martin2025pnpflow} denoisers can effectively regularize the restoration process. This paradigm has expanded beyond denoisers to incorporate diverse restoration models~\cite{hu2024a, terris2025fire}. Driven by advances in generative modeling, recent frameworks integrates GANs~\cite{goodfellow_generative_2014}, VAEs~\cite{kingma2013auto}, normalizing flows~\cite{dinh2017density}, diffusion~\cite{DeepUnsupervisedLearning2015sohl-dickstein, DenoisingDiffusionProbabilistic2020hoa, DenoisingDiffusionImplicit2020song, song2021scorebased}, and flow models~\cite{liu2023flow, lipman2023flow, albergo2025stochastic} as powerful image priors.

\begin{wrapfigure}{R}{0.5\textwidth}
  \centering 
  \vspace{-2em}
  \begin{tikzpicture}
    \node[anchor=south west, inner sep=0] (celebaplot) at (0,0) {\includegraphics[width=\linewidth]{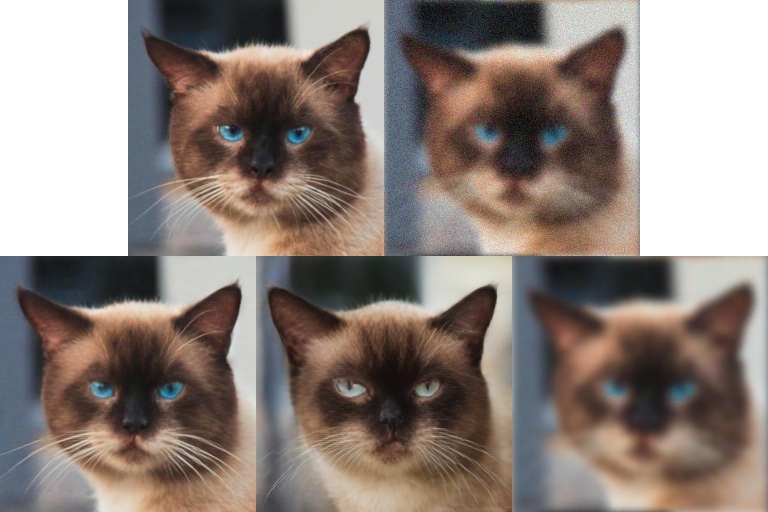}};
    
    \begin{scope}[x={(celebaplot.south east)}, y={(celebaplot.north west)}]
        \node [font=\scriptsize] at (0.33, 1.03) {Clean};
        \node [font=\scriptsize] at (0.67, 1.03) {Degraded};
        \node [font=\scriptsize] at (0.167, -0.05) {\coolname};
        \node [font=\scriptsize] at (0.5, -0.05) {S-PGD};
        \node [font=\scriptsize] at (0.833, -0.05) {S-GD};
    \end{scope}
  \end{tikzpicture}
  \captionof{figure}{\textbf{SE baseline qualitative comparisons for AFHQ deblurring}. While S-GD fails to fully project onto the natural image manifold, and S-PGD struggles to remain faithful to the original data, \coolname successfully balances both constraints to restore a sharp, high-fidelity image.}
  \label{fig:method_baselines}
  \vspace{-3em}
\end{wrapfigure}

A common direction to solve a MAP problem in a generative model’s latent space~\cite{bora2017compressed, pan2021exploiting, asim2020invertible, menon2020pulse, whang2021solving, prost2023inverse}, requires optimizing the latent vector such that applying the model and the degradation operator $A$, will result in an image similar to the observed measurement, $y$. The results of these methods are often poor, as they require propagating gradients through the generative model. 

This approach is not easily applied to diffusion/flow models due to their iterative nature and non-trivial latent space.
Nevertheless, many methods were developed to apply the image prior learned by diffusion/flow models for zero-shot restoration, often by combining the diffusion model and the acquisition process, via Bayes' rule~\cite{DenoisingDiffusionRestoration2022kawara, DiffusionPosteriorSampling2022chunga, ZeroShotImageRestoration2022wangb, PseudoinverseGuidedDiffusionModels2022songa, SolvingLinearInverserout2023, PGDiffGuidingDiffusion2023yang, DifFaceBlindFace2023yue,  pokle2024trainingfree, ben2024d, zhang2024flow, Raphaeli_2025_ICCV, man2025elad, garber2025zero}.
Their reliance on gradients computed during inference, amalgamated with their iterative nature, results in high-quality restorations at the cost of slow inference.

Evaluating image restoration methods requires using perception (whether a reconstructed image looks natural) and distortion (whether a reconstructed image resembles the unknown original image) metrics. It is well established that there is an inherent tradeoff between these metrics~\cite{PerceptionDistortionTradeoff2018blaua, TheoryDistortionPerceptionTradeoff2021freirich}, therefore, many methods excel at either of these metrics, while sacrificing the other. 

In this work, we utilize the recent SE~\cite{yue2026image} for Plug-and-Play/zero-shot image restoration. 
In contrast to previous approaches that relied on propagated gradients through the generative model, we use a fast, gradient-free iterative process that takes advantage of SE strong generative prior.

%% file: sections/4_method.tex
\section{Method}
\label{sec:method}

To achieve a MAP estimator, we need to obtain a prior over natural images, to guide the restoration process towards a plausible-looking image. Additionally, a measurement-enforcing term is needed to ground the solution in the known, available data.

In this section, we first define the MAP objective within the SE latent space and evaluate two baseline optimization strategies that rely exclusively on the decoder. Next, we analyze the Spherical Encoder’s behavior as a projection operator, highlighting its unique stability and manifold-mapping properties in comparison to standard VAEs. Building on these insights, we introduce \coolname: a simple, fast, and effective PnP algorithm. By embedding the SE projection within a Half-Quadratic Splitting (HQS) framework, \coolname alternates between manifold refinement and proximal data-consistency updates to achieve high-fidelity restoration without requiring expensive gradient backpropagation through the decoder or encoder.

\subsection{Naive MAP with a sphere prior}

Given a pretrained spherical prior, we can formulate the restoration task as a MAP estimation problem:
\begin{equation}\label{eq:proximal_operator_map}
    \underset{x}{\arg\max} \Big\{ p(x\vert y)\Big\} = \underset{x}{\arg\min} \Big\{ -\log{p(y\vert x)} - \log{p(x)} \Big\},
\end{equation}
where the right term results from applying Bayes' theorem and taking the negative logarithm.

Assuming that the decoder $D$ is injective, such that every distinct latent vector $v$ maps to a unique image $x$, we can reparameterize the problem over the latent space $\mathcal{V}$ rather than the image space $\mathcal{X}$. Furthermore, following the assumption of additive Gaussian noise in the measurements as defined in \cref{eq:degrdation}, the objective becomes:
\begin{equation}\label{eq:proximal_operator_latent}
    \underset{v}{\arg\min} \Big\{ \lVert AD(v)-y \rVert_2^2 - \log{p(v)} \Big\}.
\end{equation}
This formulation permits us to treat the prior term independently. 

\begin{figure}
  \centering

  \includegraphics[width=\linewidth]{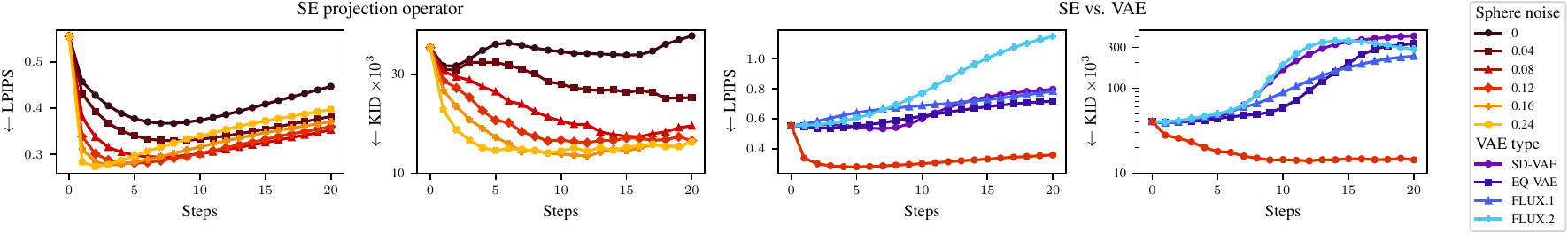}

  \caption{\textbf{SE as a projection operator to the clean image manifold.} 
  In both panels, the left plot shows LPIPS, and the right shows KID (on a log scale). Step 0 shows the results of the blurred, noisy input before projections.
  \textit{(left)} Performance of iteratively applying SE as a projection operator for different values of latent noise. SE projects degraded images to the image manifold, as illustrated by the decrease in the KID score with consecutive steps. 
  \textit{(right)} Different state-of-the-art VAEs are poor projection operators. SE is illustrated with latent noise of 0.12 on the right.}
  \label{fig:motivation}
\end{figure}

We consider two baseline solutions for handling the spherical prior. The first approach strictly enforces the prior by assuming the distribution is uniform over the shell of a sphere $\mathcal{S} = \{v \in \mathcal{V} \mid \|v\|_2^2 = L\}$ and zero elsewhere, $p(v) \propto \delta(\lVert v \rVert_2^2-L)$, allowing us to define $I_\mathcal{S}(v)$ as an indicator function that evaluates to $0$ if $v \in \mathcal{S}$ and $\infty$ otherwise. 
This leads to a constrained optimization problem:
\begin{equation}\label{eq:hard_opt}
\begin{aligned}
    &\underset{v}{\arg\min}
    && \Big\{ \lVert AD(v)-y \rVert_2^2 \Big\} \quad \text{s.t.} \quad I_\mathcal{S}(v)=0,
\end{aligned}
\end{equation}
While this is intractable, we approach a local solution using projected gradient descent (S-PGD), where in each iteration the gradient is projected onto the tangent plane to $v$, and the updated latent vector is retracted back onto the sphere to ensure a valid solution.

An alternative approach that facilitates standard gradient descent (S-GD) involves a ``soft'' enforcement of the prior. Here, we assume the distribution is Gaussian with its mean concentrated on the shell of the sphere, yielding:

\begin{equation}\label{eq:soft_opt}
\begin{aligned}
    &\underset{v}{\arg\min}
    && \Big\{ \lVert AD(v)-y \rVert_2^2 + \lambda \big\lVert \lVert v \rVert_2^2-L \big\rVert_2^2 \Big\}.
\end{aligned}
\end{equation}

Notably, in both \cref{eq:hard_opt,eq:soft_opt}, the encoder is entirely absent; gradients are propagated only through the decoder. 
This is an unfortunate omission, as the encoder is an integral part of the prior, having been trained jointly with the decoder to map images into the latent manifold~\cite{yue2026image}. Our proposed method, \coolname, utilizes both the encoder and the decoder to improve restoration performance, as we discuss in the next subsections.

Shown qualitatively in \cref{fig:method_baselines}, S-GD fails to recover sharp details, while S-PGD tends to deviate from the measurement constraints. In contrast, \coolname~produces high-fidelity details while remaining faithful to the observed data. Quantitative results in \cref{tab:baselines} further demonstrate that \coolname~outperforms both S-GD and S-PGD across PSNR, LPIPS~\cite{UnreasonableEffectivenessDeep2018zhanga}, KID~\cite{DemystifyingMMDGANs2018binkowski}, and computational runtime.

\subsection{Spherical Encoder as projection operator}\label{sec:method_se_projection}

As established, formulating the optimization exclusively over the decoder space discards the rich manifold information captured by the encoder. To fully utilize the pretrained spherical prior, we introduce a projection operator that leverages the complete autoencoder architecture. 

Denote by $P(\cdot)\,{=}\,D\left( S\left( E\left( \cdot \right), \sigma \right) \right)$ the projection operator onto the image manifold using an SE and latent random noise of strength $\sigma$. Such an operator holds several unique properties that make it attractive for image restoration: (1) $P$ projects degraded, out-of-distribution (OOD), images to semantically similar natural images; (2) each application of $P$ produces a probable sample from the image manifold; and (3) successive applications further refine the sample's quality.

We demonstrate those properties on the left side of \cref{fig:motivation}: We take 100 images from the AFHQ-Cat dataset and degrade them using Gaussian blur and additive Gaussian noise. On the degraded inputs, we repeatedly apply $P$ with varying amounts of latent noise. We evaluate the results using LPIPS to measure distortion relative to the clean version of each input and KID to measure perceptual quality relative to the entire dataset.

Empirically, the initial projection validates our first claim: within the first few steps, both LPIPS and KID significantly drop from their starting values across all evaluated noise levels. This indicates a rapid and effective mapping of OOD inputs back toward the natural image distribution. 

However, the inclusion of latent noise (${\sigma\,{>}\,0}$) within $S$ is critical to satisfying our second claim. As observed in the KID trajectories, a purely deterministic projection ($\sigma\,{=}\,0$) slowly diverge, and its perceptual quality remains noticeably worse than configurations with injected latent noise. These higher noise levels effectively stabilize the operator, ensuring it consistently produces more probable, realistic samples from the actual image manifold.

Successive applications of $P$ further validate our third claim. 
Unconstrained, repeated projections push the sample deeper into the prior manifold, refining global perceptual quality (illustrated by a stable KID score), but inevitably drift from the clean images which created the degraded inputs (illustrated by the increase in LPIPS score).

Crucially, the ability to stably apply $P$ in this manner is not a universal property of autoencoders. On the right side of \cref{fig:motivation}, we compare the SE projection against popular VAEs~\cite{rombach_high-resolution_2022, kouzelis2025eq, labs2025flux, bfl2025representation}. Instead of refining the sample, successive applications of encoding and decoding using VAEs cause both the LPIPS distortion and the KID perceptual metric to increase dramatically within just a few steps, effectively pushing the sample far from the natural image manifold. This stark contrast highlights the uniqueness of the SE: its tightly structured, bounded latent space, prevents such catastrophic divergence, uniquely enabling the stable, iterative behavior necessary in our case.

This eventual rise in distortion during unconstrained projection motivates our proposed method. 
We embed $P$ within an alternating PnP framework by interleaving the unconstrained manifold projection $P$ with a rigorous data-consistency step. This way, we harness $P$ effectiveness without sacrificing data fidelity. Our method, \coolname, effectively balances perceptual realism with adherence to the physical measurement constraints defined in \cref{eq:degrdation}.

\subsection{\coolname}

\begin{figure}[t]
  \centering
  \begin{minipage}[t]{0.52\linewidth}
    \vspace{0pt} 
    \begin{algorithm}[H]
        \small
        \caption{SP$^3$: Spherical Priors for Plug-and-Play Restoration}
        \label{alg:encdec}
        \KwInput{
        $y$; $A$; $E$; $D$;
        $S$; $\operatorname{prox}_{\lambda}$;
        $K$; $\sigma$
        }
        \KwOutput{restored image $x_K$}
        
        $x_0 \leftarrow \text{initialization from } y$\;
        
        \For{$k \leftarrow 1$ \KwTo $K$}{
            $v_k \leftarrow E(x_{k-1})$ \tcp*[r]{Encode}
            $v_k \leftarrow S(v_k, \sigma)$ \tcp*[r]{Noisy spherify}
            $x_{\mathrm{prior}} \leftarrow D(v_k)$ \tcp*[r]{Decode}
            $x_k \leftarrow \operatorname{prox}_{\lambda}(x_{\mathrm{prior}})$ \tcp*[r]{Data proximal}
        }
        \Return{$x_K$}\;
    \end{algorithm}
  \end{minipage}
  \hfill
  \begin{minipage}[t]{0.45\linewidth}
    \vspace{0pt}
    \centering
    \includegraphics[width=\linewidth]{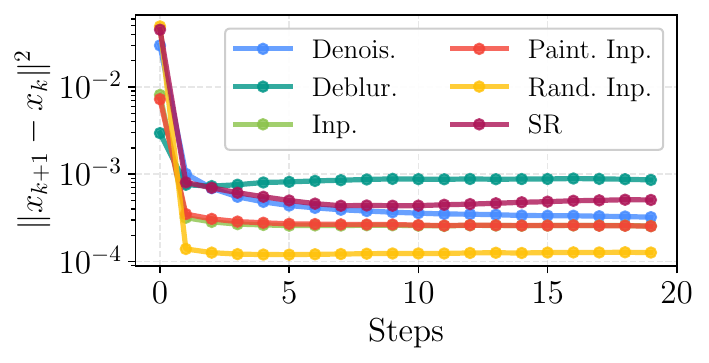}
    \captionof{figure}{\textbf{\coolname~empirical convergence.} MSE of consecutive iterations converges to zero. AFHQ Cat, y-axis is given in log-scale.}
    \label{fig:residual}
  \end{minipage}
\end{figure}

In previous sections it became evident that the naive baseline solutions, S-PGD and S-GD, yield suboptimal results. Using only the decoder, these methods fail to fully exploit the learned representations of the spherical prior. Disregarding the encoder during the restoration process effectively discards a crucial component. 

To address this, we approach the MAP estimation problem from an alternative perspective. Utilizing $I_\mathcal{S}(v)$, the strictly constrained optimization problem in \cref{eq:hard_opt} can be equivalently rewritten as an unconstrained objective:
\begin{equation}\label{eq:map_indicator}
    \underset{v}{\arg\min} \Big\{ \lVert AD(v)-y \rVert_2^2 + I_\mathcal{S}(v) \Big\}.
\end{equation}
To decouple the forward measurement model from the non-linear generative prior, we introduce an auxiliary image-space variable $x$ and enforce the constraint $x\,{=}\,D(v)$. Therefore:
\begin{equation}\label{eq:map_split}
    \underset{v,x}{\arg\min} \Big\{ \lVert Ax-y \rVert_2^2 + I_\mathcal{S}(v) \Big\} \quad \text{s.t.} \quad x=D(v).
\end{equation}
Following the Half-Quadratic Splitting (HQS) method, we relax this exact constraint by introducing a quadratic penalty term parameterized by $\lambda > 0$. This yields the unconstrained penalty function:
\begin{equation}\label{eq:hqs_penalty}
    \mathcal{L}_\lambda(x, v) = \lVert Ax-y \rVert_2^2 + I_\mathcal{S}(v) + \lambda \lVert x - D(v) \rVert_2^2.
\end{equation}

We minimize $\mathcal{L}_\lambda(x, v)$ by alternating between updating the latent representation $v$ and the image variable $x$. Given a fixed intermediate state $x_{k-1}$ from the previous iteration, updating the latent representation requires solving the following subproblem:
\begin{equation}\label{eq:hqs_prior}
    v_k = \underset{v}{\arg\min} \Big\{ I_\mathcal{S}(v) + \lambda \lVert x_{k-1} - D(v) \rVert_2^2 \Big\}.
\end{equation}
Solving \cref{eq:hqs_prior} exactly is intractable. The objective combines a non-smooth indicator function $I_S(v)$ with a non-linear decoder $D$, which precludes standard gradient-based optimization.
Instead of solving this directly, we approximate the minimizer by accounting for the SE training objectives. During training, the loss function defined in \cref{eq:se} encourages the encoder to behave such that $x \approx P(x)$, while forcing the latent to be on the learned manifold. Therefore, we bypass the optimization in \cref{eq:hqs_prior} and replace it directly with our projection operator $P(\cdot)$ introduced in \cref{sec:method_se_projection}. 

By applying this operator, the encoder-decoder pair acts as an efficient PnP projector that maps the unregularized state $x_{k-1}$ to a prior-consistent state:
\begin{equation}\label{eq:x_prior}
    x_{\mathrm{prior}} = P(x_{k-1}).
\end{equation}
Given this fixed manifold projection $x_{\mathrm{prior}}$, the alternating update for $x$ becomes a standard regularized inverse problem:
\begin{equation}\label{eq:hqs_data}
    x_k = \underset{x}{\arg\min} \Big\{ \lVert Ax-y \rVert_2^2 + \lambda \lVert x - x_{\mathrm{prior}} \rVert_2^2 \Big\}.
\end{equation}
Because $A$ is a linear operator, \cref{eq:hqs_data} represents a quadratic data-consistency step with a closed-form proximal solution:
\begin{equation}\label{eq:closed_form_x}
    x_k = \text{prox}_{\lambda}(x_{\mathrm{prior}})=(A^\top A + \lambda I)^{-1}(A^\top y + \lambda x_{\mathrm{prior}}).
\end{equation}
Thus, each iteration of our method alternates between ensuring data consistency with the measurements and explicitly projecting the result onto the learned image manifold. This formulation gives rise to our simple and efficient restoration algorithm, summarized in \cref{alg:encdec}.
Note that during the algorithm, differentiation is not required, thereby significantly reducing computational time.

To conclude this section, we discuss the convergence behavior of \coolname. By construction, our alternating update scheme is an instance of HQS. In classical optimization, if the projection operator $P(x)$ perfectly solves the prior proximal subproblem, HQS is guaranteed to converge to a stationary point of the MAP objective, even in non-convex settings~\cite{10.1007/s10107-013-0701-9, doi:10.1137/030600862}. 

In the context of PnP methods, theoretically guaranteeing this convergence requires treating the update step as a fixed-point iteration, $x_{k+1} = T(x_k) = \text{prox}_{\lambda}(P(x_k))$ \cite{hurault2022gradient}. Formal convergence to a fixed point $x_\infty \in \operatorname{Fix}(T)$ requires strong assumptions regarding the contractive or averaged nature of the projection operator $P(x)$ (see \cref{app:theoretical analysis} for an extended fixed-point theorem and proofs). Because guaranteeing these properties for highly non-linear generative priors is restrictively strong in practice, we validate the convergence of \coolname~empirically. As illustrated in \Cref{fig:residual}, the empirical sequence generated by our method rapidly stabilizes. Across all evaluated restoration tasks on the AFHQ-Cat dataset, the residual converges to near-zero within the first few iterations, demonstrating that \coolname robustly reaches a stable fixed point on the image manifold.

%% file: sections/5_experiments.tex
\section{Experiments}
\label{sec:experiments}

\begin{figure}
  \centering

  \includegraphics[width=\linewidth]{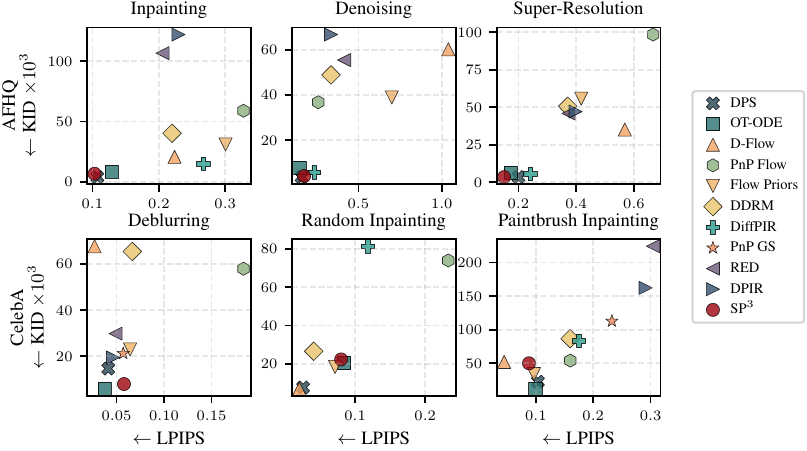}
    
  \caption{\textbf{Perception-distortion evaluation.} Comparison between \coolname~(red sphere) and competing methods on perception (KID) and distortion (LPIPS) metrics on AFHQ (top) and CelebA (bottom). \coolname~achieves at least comparable perception-distortion results while being significantly faster.}
  \label{fig:metrics}
\end{figure}

We compare \coolname with zero-shot image restoration methods. The compared methods are divided into three categories: (1) PnP methods utilizing neural denoisers~\cite{LittleEngineThat2017romanoa, PlugPlayImageRestoration2022zhang, hurault2022gradient}; (2) PnP methods utilizing generative (diffusion/flow based) denoisers~\cite{ DenoisingDiffusionModels2023zhua, martin2025pnpflow}; and (3) diffusion/flow based zero-shot methods~\cite{DenoisingDiffusionRestoration2022kawara, DiffusionPosteriorSampling2022chunga, pokle2024trainingfree,zhang2024flow,ben2024d}.
Following the recent PnP-Flow~\cite{martin2025pnpflow}, we test the different methods on both AFHQ-Cat~\cite{choi2020AFHQ} at 256px resolution and CelebA~\cite{yang2015facial} at 128px resolution. We use the test sets defined by \cite{martin2025pnpflow}.
All the results describe in this section are for 20 \coolname steps, unless specifically mentioned otherwise.
Below we provide details about the experimental setting, where additional implementation details are provided in \cref{app:implementation}.

\begin{figure}[t]
    \centering
    \begin{tikzpicture}
        \node[anchor=south west, inner sep=0] (top_cat) at (0,5.23) {\includegraphics[width=\linewidth]{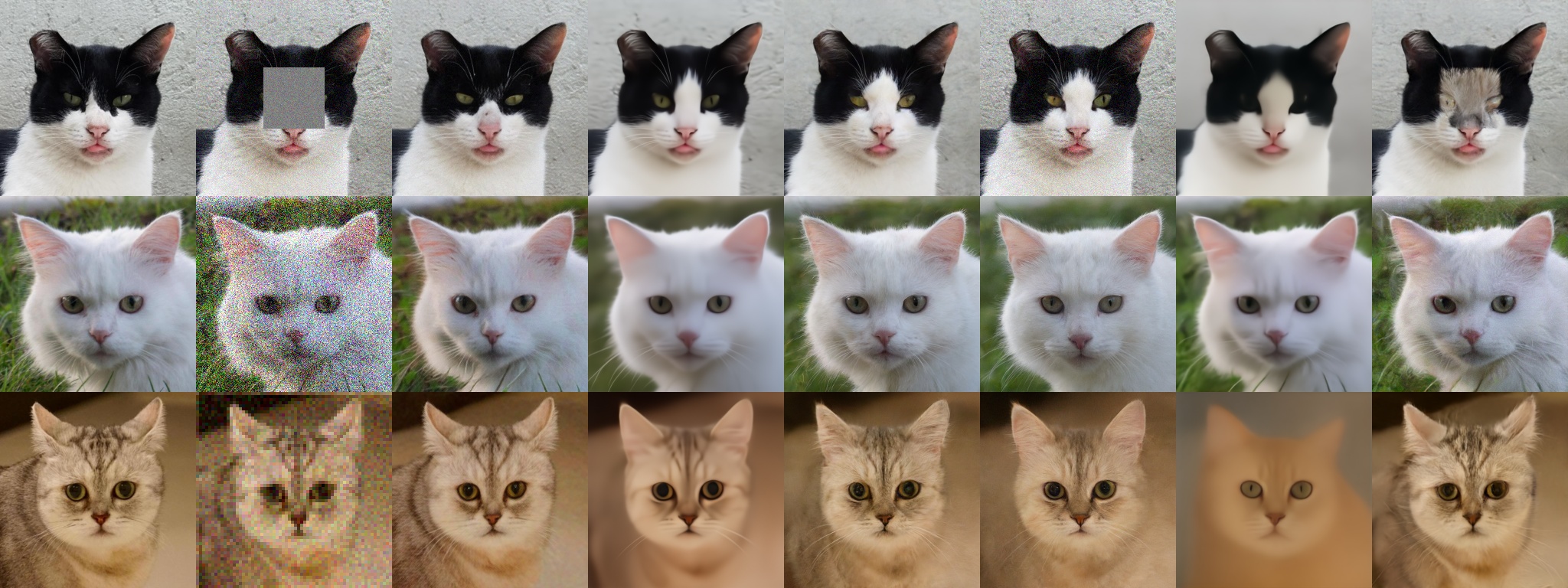}};
        
        \node[anchor=south west, inner sep=0] (bottom_cat) at (0,0) {\includegraphics[width=\linewidth]{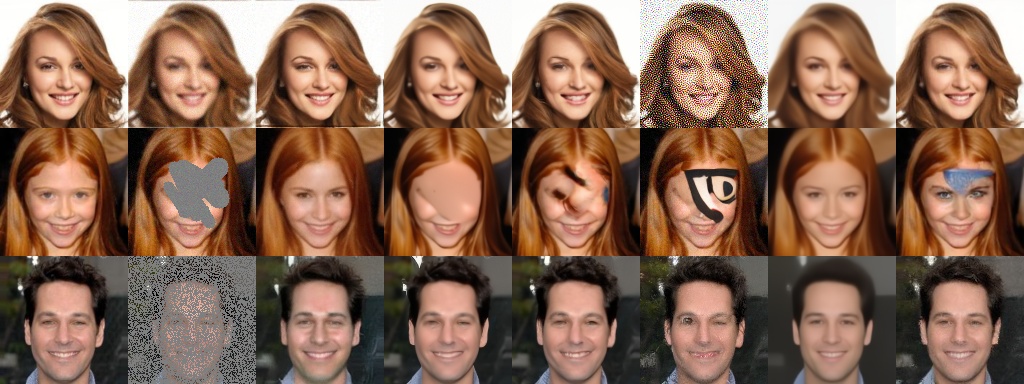}};
        
        \begin{scope}[shift={(top_cat.north west)}, x=\linewidth]
            \tikzset{every node/.style={anchor=base, font=\small, text height=1.5ex, text depth=.25ex}}
            
            \node at (0.0625, 0.05) {Clean};
            \node at (0.1875, 0.05) {Degraded};
            \node at (0.3125, 0.05) {\coolname{}};
            \node at (0.4375, 0.05) {DDRM};
            \node at (0.5625, 0.05) {DPS};
            \node at (0.6875, 0.05) {DiffPIR};
            \node at (0.8125, 0.05) {PnP-Flow};
            \node at (0.9375, 0.05) {OT-ODE};
        \end{scope}
    \end{tikzpicture}
    \caption{\textbf{Qualitative comparison of image restoration.} Results for AFHQ-Cat (top rows) and CelebA (bottom rows). The six rows correspond to the following degradation in descending order: inpainting, denoising, super-resolution, deblurring, paintbrush-inpainting, and random inpainting.}
    \label{fig:quality_comparisons}
\end{figure}

\paragraph{Priors.} For methods utilizing flow models, we use the priors trained by PnP-Flow~\cite{martin2025pnpflow}. For diffusion-based methods, we use publicly available diffusion models~\cite{cyclediffusion, rdruce-2026-ddpm-celeb-128}. 
Classical PnP methods use the DRUnet and GS-DRUnet implemented by the DeepInv library~\cite{tachella2025deepinverse}.
For SE, we use the official code~\cite{github_sphere_encoder} to train an AFHQ-Cat model using the official hyper-parameters described in~\cite{yue2026image}, and a CelebA model using the same settings for 100 epochs.
To ensure fair comparison between SE and diffusion/flow-based methods, we validate that the generative models used as the backbone for all methods mostly achieve similar KID results for generation. See \cref{app:implementation} for additional details. 

\paragraph{Degradations.} For each dataset, we evaluate all methods across six noisy image restoration tasks, in the form of \Cref{eq:degrdation}: (1) Gaussian denoising with $\sigma=0.4$ for AFHQ-Cat and $\sigma=0.2$ for CelebA; (2) Deblurring using a $61\times 61$ Gaussian kernel with $\sigma_{\text{blur}}=3.0$ for AFHQ-Cat and $\sigma_{\text{blur}}=1.0$ for CelebA; (3) Super-resolution using a $4\times$ bicubic downsampling kernel for AFHQ-Cat and $2\times$ for CelebA; (4) Box-inpainting with a centered square mask of size $80$px for AFHQ-Cat and $40$px for CelebA; (5) Random pixel inpainting with 70\% masked pixels; and (6) paintbrush inpainting. For random inpainting, we add Gaussian noise with $\sigma=0.02$ on AFHQ-Cat and $\sigma=0.01$ on CelebA; for all other restoration tasks, we use $\sigma=0.1$ on AFHQ-Cat and $\sigma=0.05$ on CelebA.

\paragraph{Metrics.} We measure both distortion and perception metrics as they are at odds with each other~\cite{PerceptionDistortionTradeoff2018blaua, TheoryDistortionPerceptionTradeoff2021freirich}. We use LPIPS~\cite{UnreasonableEffectivenessDeep2018zhanga} as our main distortion metric, and KID~\cite{DemystifyingMMDGANs2018binkowski} as our perception index. 
In \cref{app:comparisons} we report the complete set of results in table form, including PSNR as an additional distortion metric that favors smooth reconstructions. Moreover, we report the per-image wall-clock runtime of each method on a single NVIDIA L40S GPU.

\paragraph{Comparisons.} \Cref{fig:metrics} illustrates quantitative results of \coolname~(red sphere) and the competing methods on a perception-distortion plane~\cite{PerceptionDistortionTradeoff2018blaua}. 
The top part shows results for AFHQ, and the bottom part for CelebA, with each plot showing a different degradation. \coolname achieves comparable results to those of the best competing methods, typically DPS~\cite{DiffusionPosteriorSampling2022chunga} and OT-ODE~\cite{pokle2024trainingfree}.
Qualitative comparisons illustrated in \cref{fig:quality_comparisons}, where \coolname produces sharp results with fine details that are consistent with the degraded input and loyal to the clean image. 
See~\Cref{app:comparisons} for additional visual comparisons.

\begin{figure*}[t] %
    \centering
    
    \begin{minipage}[b]{0.64\linewidth}
        \centering
        \includegraphics[width=\linewidth]{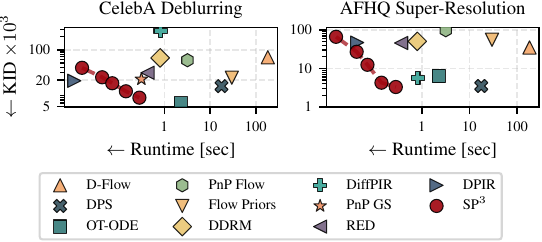}
        \caption{\textbf{Perceptual quality vs.~running time.} Comparison between \coolname~(red sphere) and competing methods on perception (KID) and running time (sec). \coolname~achieves at least comparable perceptual quality at fraction of the time, achieving $3$-$630\times $ speedup. \coolname~results are displayed for 1, 3, 5, 10, and 20 steps.}
        \label{fig:time_kid}
    \end{minipage}
    \hfill %
    \begin{minipage}[b]{0.34\linewidth}
        \centering
        \begin{tikzpicture}
            \node[anchor=south west, inner sep=0] (celebaplot) at (0,0) {\includegraphics[width=\linewidth]{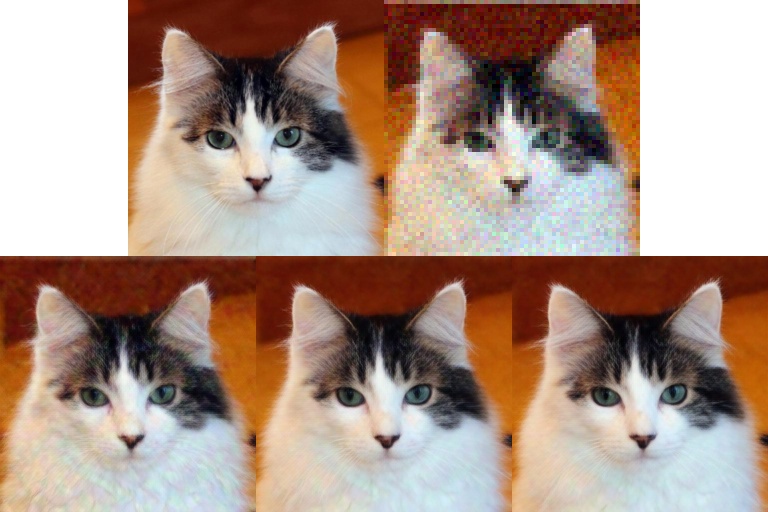}};
            
            \begin{scope}[x={(celebaplot.south east)}, y={(celebaplot.north west)}]
                
                \node [font=\scriptsize] at (0.33, 1.037) {Clean};
                \node [font=\scriptsize] at (0.67, 1.03) {Degraded};
                
                \node [font=\scriptsize] at (0.167, -0.05) {1 step};
                \node [font=\scriptsize] at (0.5, -0.05) {3 steps};
                \node [font=\scriptsize] at (0.833, -0.05) {5 steps};
                
            \end{scope}
        \end{tikzpicture}
        \caption{\textbf{Anytime results.} Results of anytime restoration for super-resolution on AFHQ.}
        \label{fig:anytime_example}
    \end{minipage}
    
\end{figure*}

\paragraph{Speed \& anytime generation.} 
\cref{fig:metrics,fig:quality_comparisons} do not reveal one of the key advantages of \coolname, its speed and anytime capabilities. 
Because SE projects degraded images onto the clean-image manifold, \coolname achieves perceptual results comparable to those of competing methods even after a single step. 
This is illustrated in \cref{fig:time_kid}, where we compare the different methods on the perception-runtime plane. Even after a single step (leftmost point), \coolname is better than several competing methods, and after 20 steps (rightmost point), \coolname is comparable to the best competing methods while being significantly faster. 
The speed advantage stems from the small number of steps required, without the need to compute gradients. Concretely, \coolname using $20$ steps is $\bm{\times8}$ faster than OT-ODE and $\bm{\times60}$ faster than DPS, which are its most direct competitors in terms of visual quality as seen in \cref{fig:metrics,fig:quality_comparisons}. 
Finally, \cref{fig:anytime_example} presents an example of the restoration trajectory, demonstrating \coolname anytime capability. Even after a single step, we obtain a usable, sharp result that can be further refined.

%% file: sections/6_conclusion.tex
\section{Conclusion and limitations}
\label{sec:conclusion}

We introduce \coolname, a fast, gradient-free Plug-and-Play algorithm that leverages Spherical Encoders to solve the MAP image restoration problem. By interleaving robust manifold projections with a closed-form data-consistency step, \coolname achieves anytime generation and perceptual quality matching state-of-the-art diffusion and flow methods, at a fraction of the computational cost.

Despite its empirical success, our method presents some limitations. As the projection operator blindly forces intermediate states back onto the clean image manifold, \coolname strongly prioritizes perceptual realism. Consequently, it naturally lacks in classical distortion metrics and cannot provide Minimum Mean Square Error (MMSE) solutions, which inherently favor smoother, averaged outputs. Additionally, while \coolname efficiently solves the MAP objective, it is not a posterior sampler like DPS~\cite{DiffusionPosteriorSampling2022chunga}. 
Investigating how the structural latent noise could be actively leveraged to sample from and explore the posterior distribution~\cite{cohen_posterior_2023, nehme_generative_2025} is a promising direction for future work. 

\paragraph{Societal impacts.} Our method restores clean images from partial measurements, benefiting applications like smartphone photography and noisy imaging. However, it poses dual-use risks, including unauthorized surveillance and deepfake generation.

%% file: sections/B_comparisons.tex
\section{Additional Comparisons}
\label{app:comparisons}

\begin{figure}[htbp]
    \centering
    \begin{tikzpicture}
        \node[anchor=south west, inner sep=0] (row1) at (0,7.7) {\includegraphics[width=\linewidth]{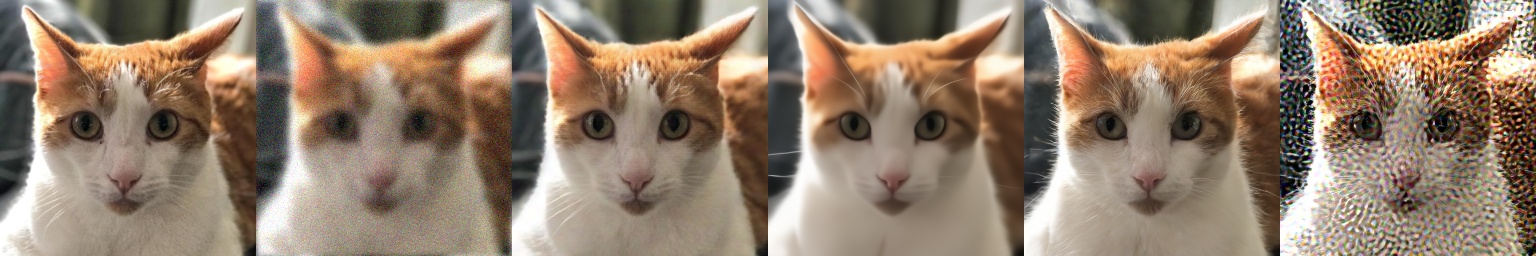}};
        \begin{scope}[shift={(row1.north west)}, x=\linewidth]
            \tikzset{every node/.style={anchor=base, font=\scriptsize, text height=1.5ex, text depth=.25ex}}
            \node at (0.083, 0.05) {Clean};
            \node at (0.250, 0.05) {Degraded};
            \node at (0.416, 0.05) {\coolname{}};
            \node at (0.583, 0.05) {DDRM};
            \node at (0.750, 0.05) {DPS};
            \node at (0.916, 0.05) {PnP-Diff};
        \end{scope}

        \node[anchor=south west, inner sep=0] (row2) at (0,5.37) {\includegraphics[width=\linewidth]{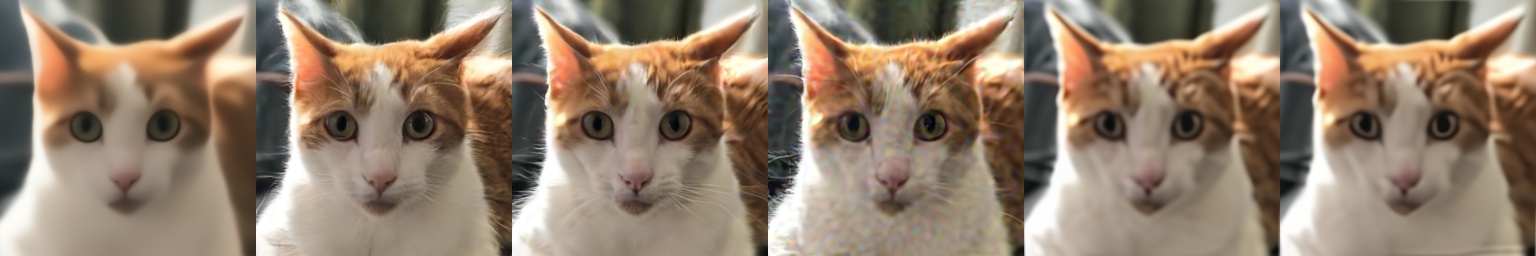}};
        \begin{scope}[shift={(row2.south west)}, x=\linewidth]
            \tikzset{every node/.style={anchor=base, font=\scriptsize, text height=1.5ex, text depth=.25ex}}
            \node at (0.083, -0.2) {PnP-Flow};
            \node at (0.250, -0.2) {OT-ODE};
            \node at (0.416, -0.2) {D-Flow};
            \node at (0.583, -0.2) {Flow Priors};
            \node at (0.750, -0.2) {DPIR};
            \node at (0.916, -0.2) {RED};
        \end{scope}

        \node[anchor=south west, inner sep=0] (row3) at (0,2.33) {\includegraphics[width=\linewidth]{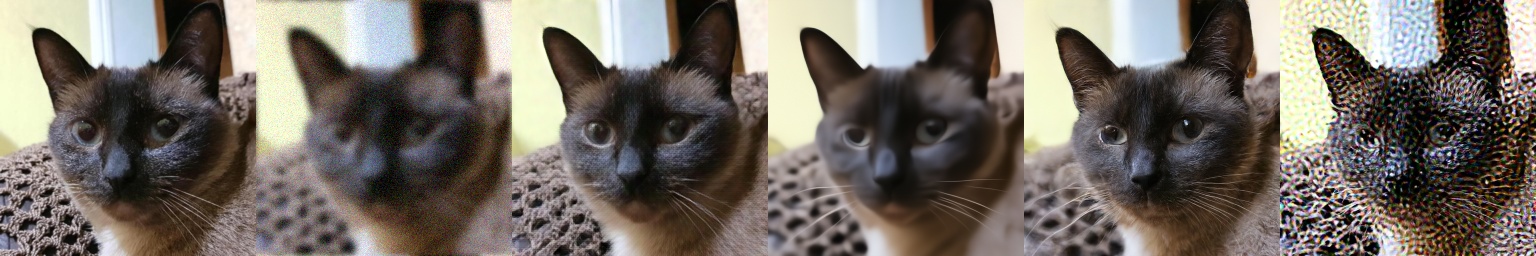}};
        \begin{scope}[shift={(row3.north west)}, x=\linewidth]
            \tikzset{every node/.style={anchor=base, font=\scriptsize, text height=1.5ex, text depth=.25ex}}
            \node at (0.083, 0.05) {Clean};
            \node at (0.250, 0.05) {Degraded};
            \node at (0.416, 0.05) {\coolname{}};
            \node at (0.583, 0.05) {DDRM};
            \node at (0.750, 0.05) {DPS};
            \node at (0.916, 0.05) {PnP-Diff};
        \end{scope}

        \node[anchor=south west, inner sep=0] (row4) at (0,0) {\includegraphics[width=\linewidth]{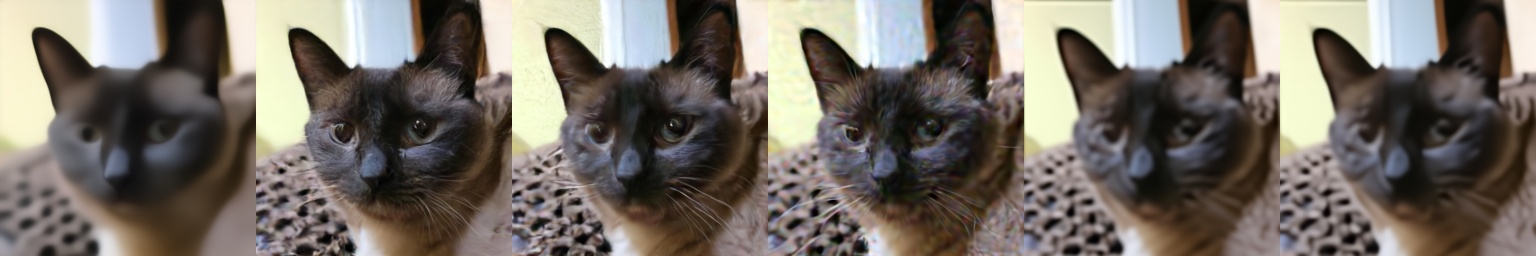}};
        \begin{scope}[shift={(row4.south west)}, x=\linewidth]
            \tikzset{every node/.style={anchor=base, font=\scriptsize, text height=1.5ex, text depth=.25ex}}
            \node at (0.083, -0.2) {PnP-Flow};
            \node at (0.250, -0.2) {OT-ODE};
            \node at (0.416, -0.2) {D-Flow};
            \node at (0.583, -0.2) {Flow Priors};
            \node at (0.750, -0.2) {DPIR};
            \node at (0.916, -0.2) {RED};
        \end{scope}
    \end{tikzpicture}
    \caption{Qualitative comparison of image restoration results on AFHQ-Cat (deblurring).}
    \label{fig:quality_comparisons_app_afhq_deblurring}
\end{figure}

\begin{figure}[htbp]
    \centering
    \begin{tikzpicture}
        \node[anchor=south west, inner sep=0] (row1) at (0,7.7) {\includegraphics[width=\linewidth]{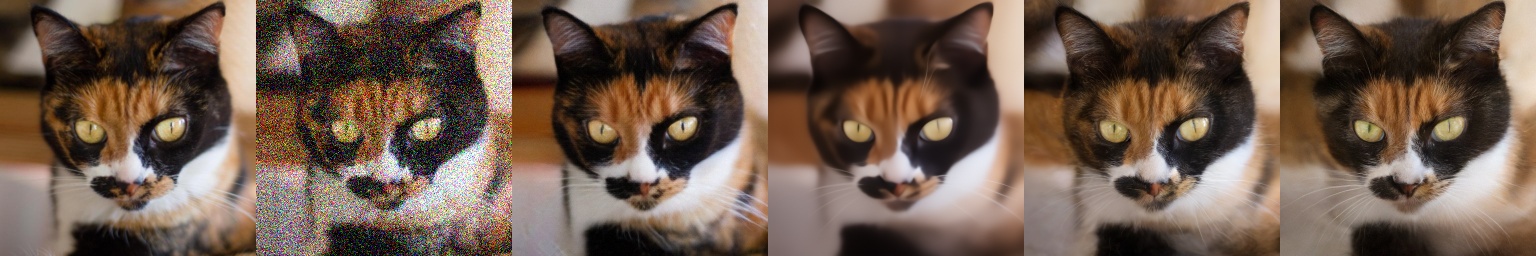}};
        \begin{scope}[shift={(row1.north west)}, x=\linewidth]
            \tikzset{every node/.style={anchor=base, font=\scriptsize, text height=1.5ex, text depth=.25ex}}
            \node at (0.083, 0.05) {Clean};
            \node at (0.250, 0.05) {Degraded};
            \node at (0.416, 0.05) {\coolname{}};
            \node at (0.583, 0.05) {DDRM};
            \node at (0.750, 0.05) {DPS};
            \node at (0.916, 0.05) {PnP-Diff};
        \end{scope}

        \node[anchor=south west, inner sep=0] (row2) at (0,5.37) {\includegraphics[width=\linewidth]{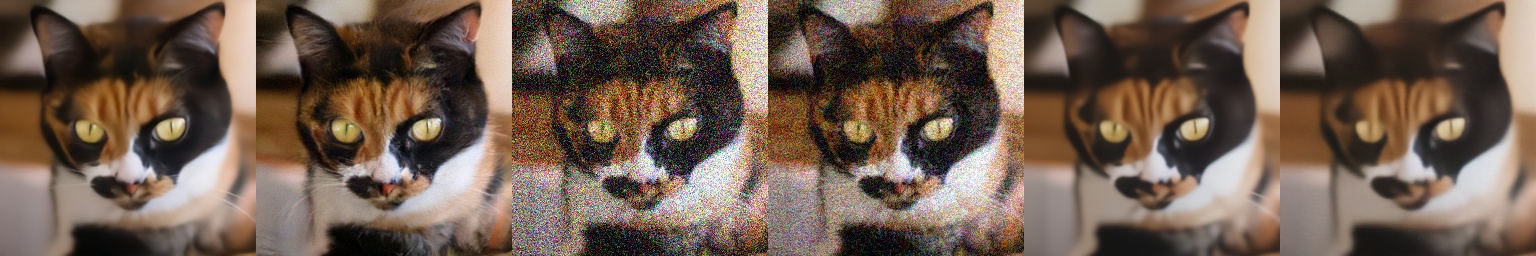}};
        \begin{scope}[shift={(row2.south west)}, x=\linewidth]
            \tikzset{every node/.style={anchor=base, font=\scriptsize, text height=1.5ex, text depth=.25ex}}
            \node at (0.083, -0.2) {PnP-Flow};
            \node at (0.250, -0.2) {OT-ODE};
            \node at (0.416, -0.2) {D-Flow};
            \node at (0.583, -0.2) {Flow Priors};
            \node at (0.750, -0.2) {DPIR};
            \node at (0.916, -0.2) {RED};
        \end{scope}

        \node[anchor=south west, inner sep=0] (row3) at (0,2.33) {\includegraphics[width=\linewidth]{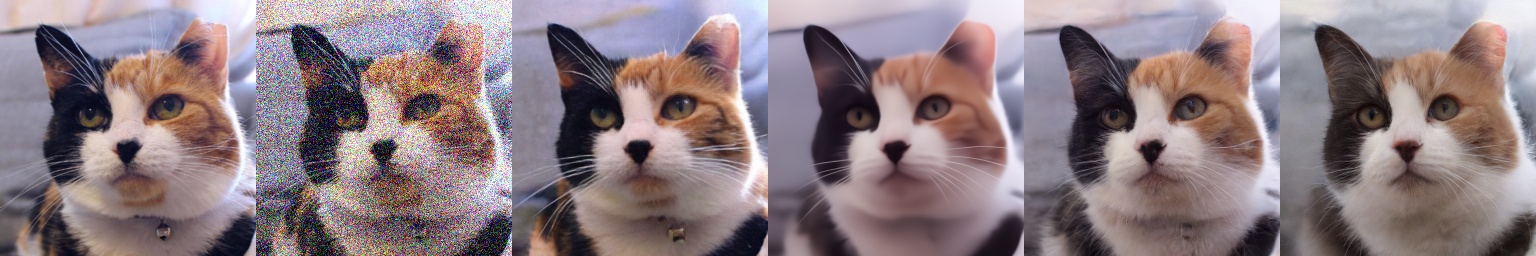}};
        \begin{scope}[shift={(row3.north west)}, x=\linewidth]
            \tikzset{every node/.style={anchor=base, font=\scriptsize, text height=1.5ex, text depth=.25ex}}
            \node at (0.083, 0.05) {Clean};
            \node at (0.250, 0.05) {Degraded};
            \node at (0.416, 0.05) {\coolname{}};
            \node at (0.583, 0.05) {DDRM};
            \node at (0.750, 0.05) {DPS};
            \node at (0.916, 0.05) {PnP-Diff};
        \end{scope}

        \node[anchor=south west, inner sep=0] (row4) at (0,0) {\includegraphics[width=\linewidth]{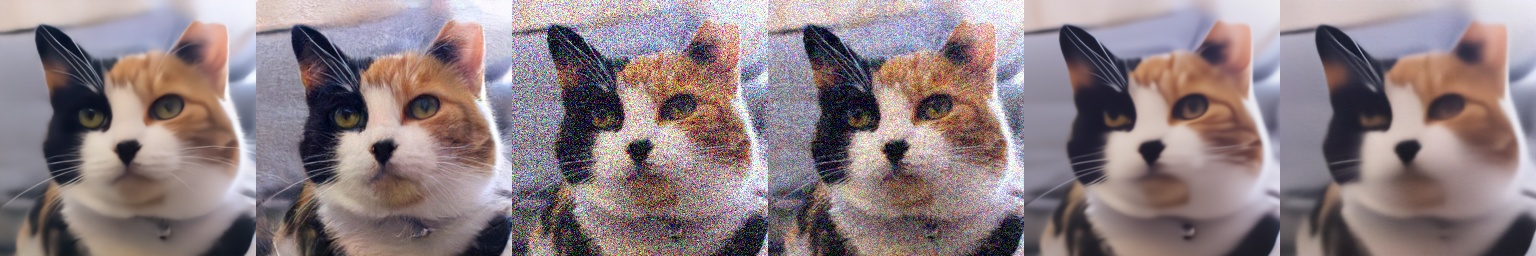}};
        \begin{scope}[shift={(row4.south west)}, x=\linewidth]
            \tikzset{every node/.style={anchor=base, font=\scriptsize, text height=1.5ex, text depth=.25ex}}
            \node at (0.083, -0.2) {PnP-Flow};
            \node at (0.250, -0.2) {OT-ODE};
            \node at (0.416, -0.2) {D-Flow};
            \node at (0.583, -0.2) {Flow Priors};
            \node at (0.750, -0.2) {DPIR};
            \node at (0.916, -0.2) {RED};
        \end{scope}
    \end{tikzpicture}
    \caption{Qualitative comparison of image restoration results on AFHQ-Cat (denoising).}
    \label{fig:quality_comparisons_app_afhq_denoising}
\end{figure}

\begin{figure}[htbp]
    \centering
    \begin{tikzpicture}
        \node[anchor=south west, inner sep=0] (row1) at (0,7.7) {\includegraphics[width=\linewidth]{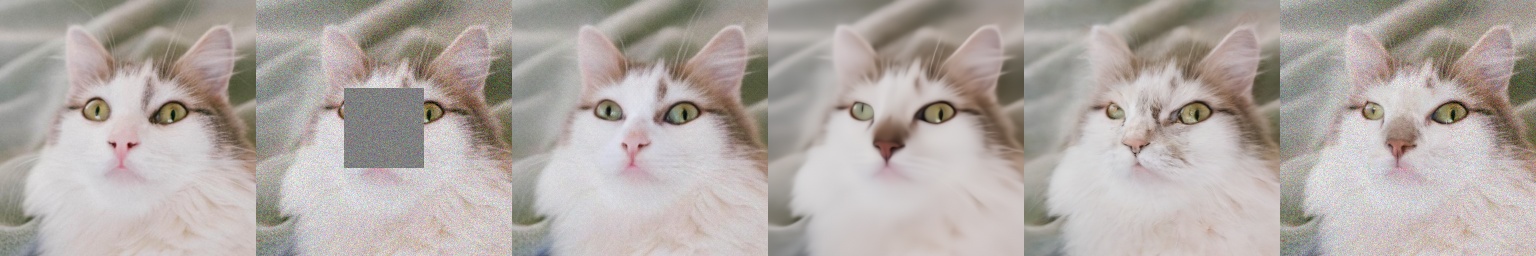}};
        \begin{scope}[shift={(row1.north west)}, x=\linewidth]
            \tikzset{every node/.style={anchor=base, font=\scriptsize, text height=1.5ex, text depth=.25ex}}
            \node at (0.083, 0.05) {Clean};
            \node at (0.250, 0.05) {Degraded};
            \node at (0.416, 0.05) {\coolname{}};
            \node at (0.583, 0.05) {DDRM};
            \node at (0.750, 0.05) {DPS};
            \node at (0.916, 0.05) {PnP-Diff};
        \end{scope}

        \node[anchor=south west, inner sep=0] (row2) at (0,5.37) {\includegraphics[width=\linewidth]{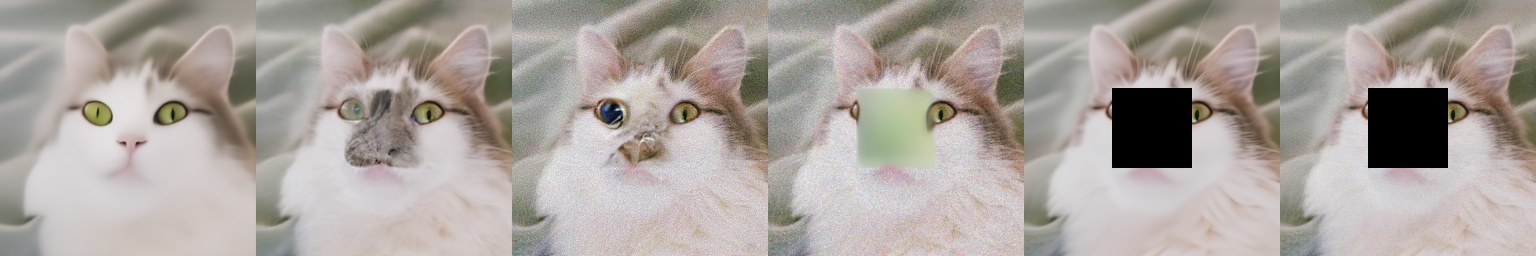}};
        \begin{scope}[shift={(row2.south west)}, x=\linewidth]
            \tikzset{every node/.style={anchor=base, font=\scriptsize, text height=1.5ex, text depth=.25ex}}
            \node at (0.083, -0.2) {PnP-Flow};
            \node at (0.250, -0.2) {OT-ODE};
            \node at (0.416, -0.2) {D-Flow};
            \node at (0.583, -0.2) {Flow Priors};
            \node at (0.750, -0.2) {DPIR};
            \node at (0.916, -0.2) {RED};
        \end{scope}

        \node[anchor=south west, inner sep=0] (row3) at (0,2.33) {\includegraphics[width=\linewidth]{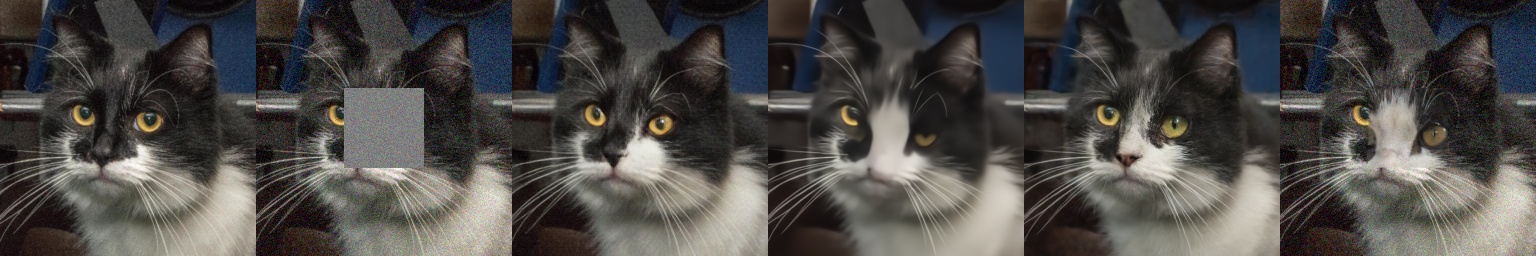}};
        \begin{scope}[shift={(row3.north west)}, x=\linewidth]
            \tikzset{every node/.style={anchor=base, font=\scriptsize, text height=1.5ex, text depth=.25ex}}
            \node at (0.083, 0.05) {Clean};
            \node at (0.250, 0.05) {Degraded};
            \node at (0.416, 0.05) {\coolname{}};
            \node at (0.583, 0.05) {DDRM};
            \node at (0.750, 0.05) {DPS};
            \node at (0.916, 0.05) {PnP-Diff};
        \end{scope}

        \node[anchor=south west, inner sep=0] (row4) at (0,0) {\includegraphics[width=\linewidth]{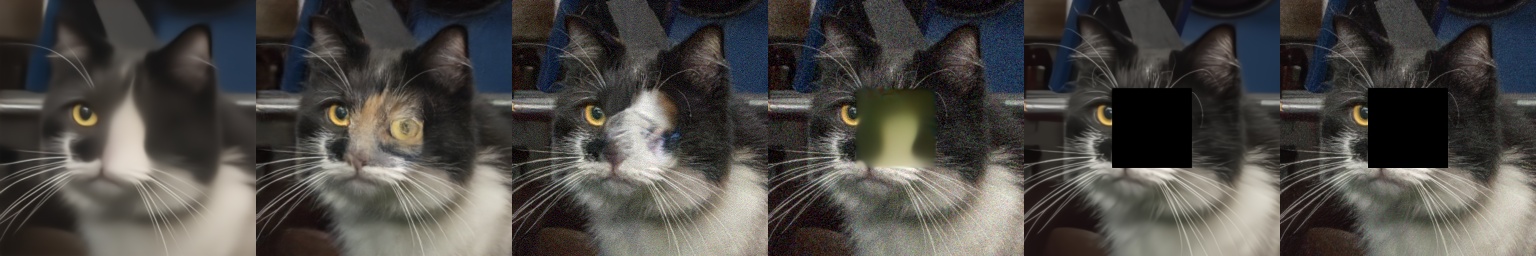}};
        \begin{scope}[shift={(row4.south west)}, x=\linewidth]
            \tikzset{every node/.style={anchor=base, font=\scriptsize, text height=1.5ex, text depth=.25ex}}
            \node at (0.083, -0.2) {PnP-Flow};
            \node at (0.250, -0.2) {OT-ODE};
            \node at (0.416, -0.2) {D-Flow};
            \node at (0.583, -0.2) {Flow Priors};
            \node at (0.750, -0.2) {DPIR};
            \node at (0.916, -0.2) {RED};
        \end{scope}
    \end{tikzpicture}
    \caption{Qualitative comparison of image restoration results on AFHQ-Cat (inpainting).}
    \label{fig:quality_comparisons_app_afhq_inpainting}
\end{figure}

\begin{figure}[htbp]
    \centering
    \begin{tikzpicture}
        \node[anchor=south west, inner sep=0] (row1) at (0,7.7) {\includegraphics[width=\linewidth]{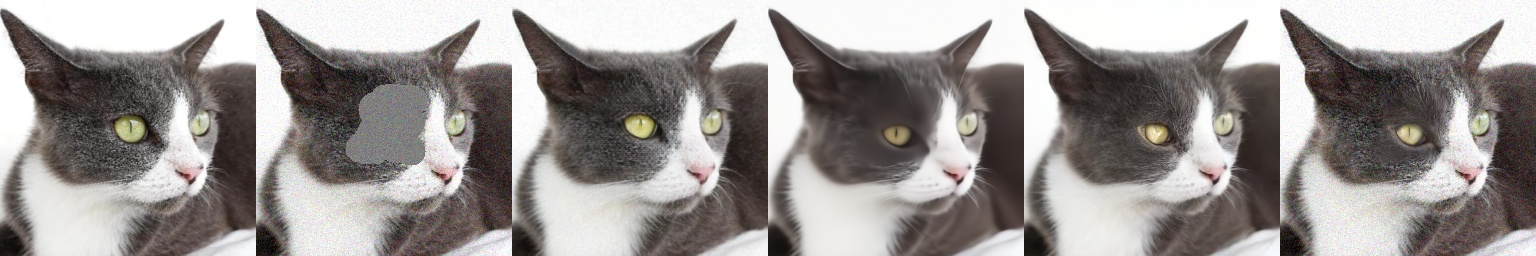}};
        \begin{scope}[shift={(row1.north west)}, x=\linewidth]
            \tikzset{every node/.style={anchor=base, font=\scriptsize, text height=1.5ex, text depth=.25ex}}
            \node at (0.083, 0.05) {Clean};
            \node at (0.250, 0.05) {Degraded};
            \node at (0.416, 0.05) {\coolname{}};
            \node at (0.583, 0.05) {DDRM};
            \node at (0.750, 0.05) {DPS};
            \node at (0.916, 0.05) {PnP-Diff};
        \end{scope}

        \node[anchor=south west, inner sep=0] (row2) at (0,5.37) {\includegraphics[width=\linewidth]{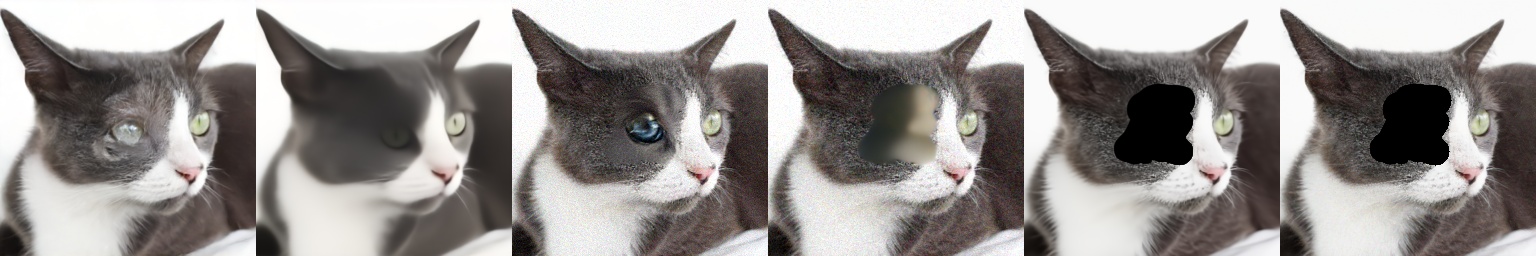}};
        \begin{scope}[shift={(row2.south west)}, x=\linewidth]
            \tikzset{every node/.style={anchor=base, font=\scriptsize, text height=1.5ex, text depth=.25ex}}
            \node at (0.083, -0.2) {PnP-Flow};
            \node at (0.250, -0.2) {OT-ODE};
            \node at (0.416, -0.2) {D-Flow};
            \node at (0.583, -0.2) {Flow Priors};
            \node at (0.750, -0.2) {DPIR};
            \node at (0.916, -0.2) {RED};
        \end{scope}

        \node[anchor=south west, inner sep=0] (row3) at (0,2.33) {\includegraphics[width=\linewidth]{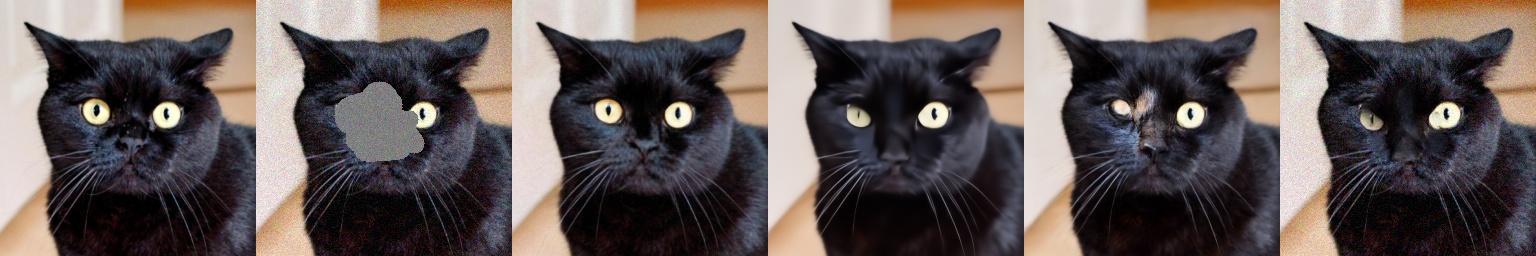}};
        \begin{scope}[shift={(row3.north west)}, x=\linewidth]
            \tikzset{every node/.style={anchor=base, font=\scriptsize, text height=1.5ex, text depth=.25ex}}
            \node at (0.083, 0.05) {Clean};
            \node at (0.250, 0.05) {Degraded};
            \node at (0.416, 0.05) {\coolname{}};
            \node at (0.583, 0.05) {DDRM};
            \node at (0.750, 0.05) {DPS};
            \node at (0.916, 0.05) {PnP-Diff};
        \end{scope}

        \node[anchor=south west, inner sep=0] (row4) at (0,0) {\includegraphics[width=\linewidth]{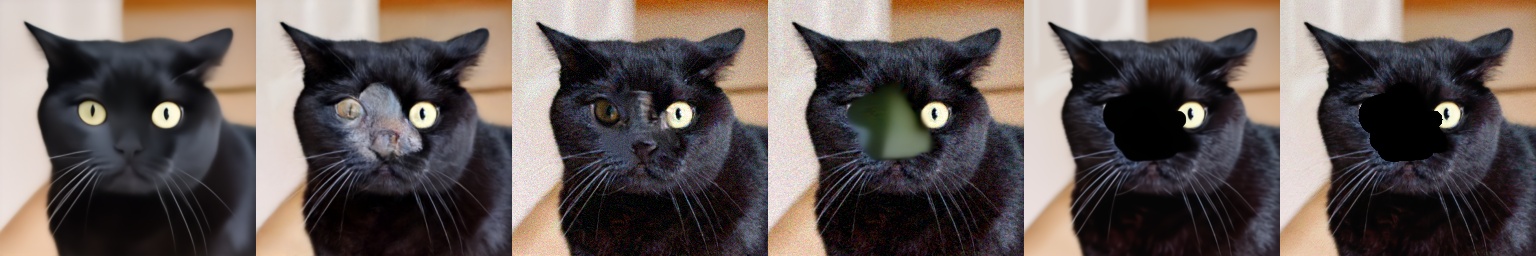}};
        \begin{scope}[shift={(row4.south west)}, x=\linewidth]
            \tikzset{every node/.style={anchor=base, font=\scriptsize, text height=1.5ex, text depth=.25ex}}
            \node at (0.083, -0.2) {PnP-Flow};
            \node at (0.250, -0.2) {OT-ODE};
            \node at (0.416, -0.2) {D-Flow};
            \node at (0.583, -0.2) {Flow Priors};
            \node at (0.750, -0.2) {DPIR};
            \node at (0.916, -0.2) {RED};
        \end{scope}
    \end{tikzpicture}
    \caption{Qualitative comparison of image restoration results on AFHQ-Cat (paintbrush-inpainting).}
    \label{fig:quality_comparisons_app_afhq_paintbrush_inpainting}
\end{figure}

\begin{figure}[htbp]
    \centering
    \begin{tikzpicture}
        \node[anchor=south west, inner sep=0] (row1) at (0,7.7) {\includegraphics[width=\linewidth]{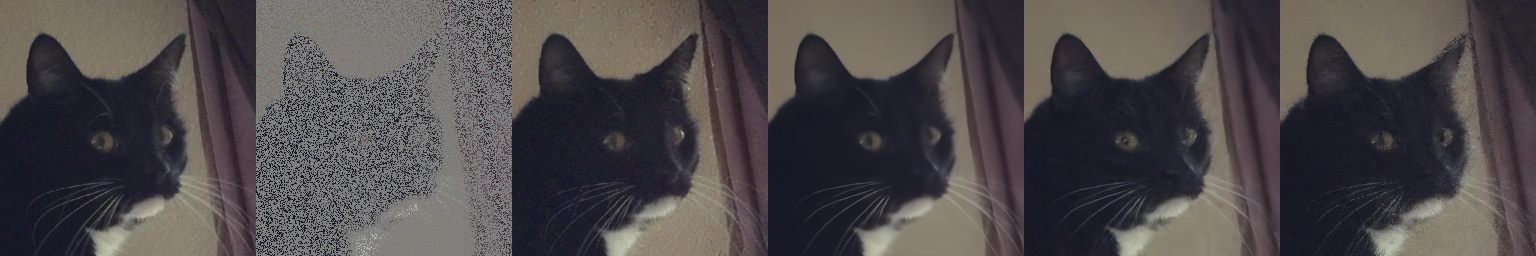}};
        \begin{scope}[shift={(row1.north west)}, x=\linewidth]
            \tikzset{every node/.style={anchor=base, font=\scriptsize, text height=1.5ex, text depth=.25ex}}
            \node at (0.083, 0.05) {Clean};
            \node at (0.250, 0.05) {Degraded};
            \node at (0.416, 0.05) {\coolname{}};
            \node at (0.583, 0.05) {DDRM};
            \node at (0.750, 0.05) {DPS};
            \node at (0.916, 0.05) {PnP-Diff};
        \end{scope}

        \node[anchor=south west, inner sep=0] (row2) at (0,5.37) {\includegraphics[width=\linewidth]{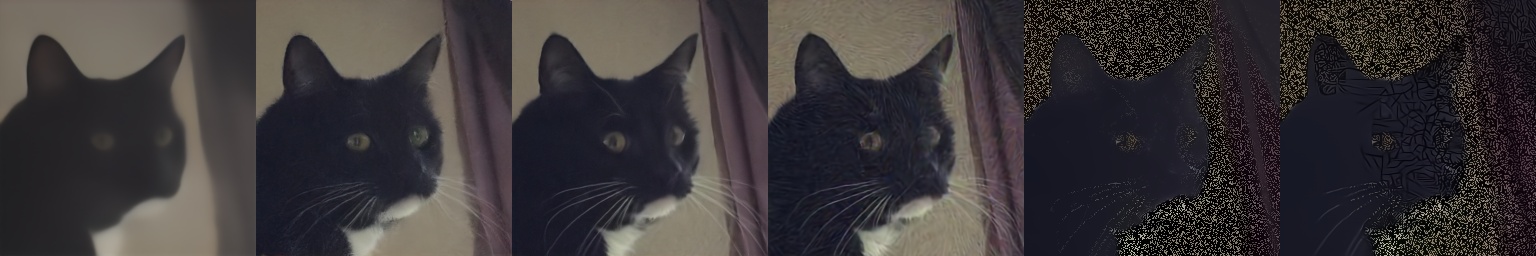}};
        \begin{scope}[shift={(row2.south west)}, x=\linewidth]
            \tikzset{every node/.style={anchor=base, font=\scriptsize, text height=1.5ex, text depth=.25ex}}
            \node at (0.083, -0.2) {PnP-Flow};
            \node at (0.250, -0.2) {OT-ODE};
            \node at (0.416, -0.2) {D-Flow};
            \node at (0.583, -0.2) {Flow Priors};
            \node at (0.750, -0.2) {DPIR};
            \node at (0.916, -0.2) {RED};
        \end{scope}

        \node[anchor=south west, inner sep=0] (row3) at (0,2.33) {\includegraphics[width=\linewidth]{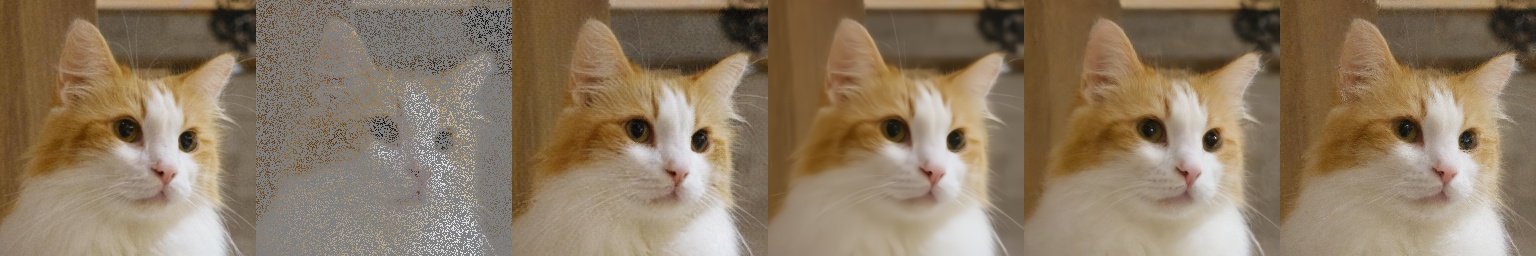}};
        \begin{scope}[shift={(row3.north west)}, x=\linewidth]
            \tikzset{every node/.style={anchor=base, font=\scriptsize, text height=1.5ex, text depth=.25ex}}
            \node at (0.083, 0.05) {Clean};
            \node at (0.250, 0.05) {Degraded};
            \node at (0.416, 0.05) {\coolname{}};
            \node at (0.583, 0.05) {DDRM};
            \node at (0.750, 0.05) {DPS};
            \node at (0.916, 0.05) {PnP-Diff};
        \end{scope}

        \node[anchor=south west, inner sep=0] (row4) at (0,0) {\includegraphics[width=\linewidth]{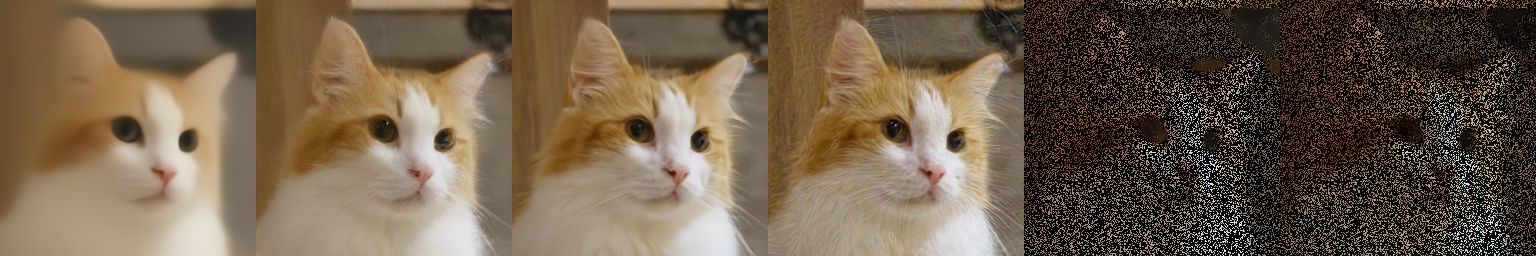}};
        \begin{scope}[shift={(row4.south west)}, x=\linewidth]
            \tikzset{every node/.style={anchor=base, font=\scriptsize, text height=1.5ex, text depth=.25ex}}
            \node at (0.083, -0.2) {PnP-Flow};
            \node at (0.250, -0.2) {OT-ODE};
            \node at (0.416, -0.2) {D-Flow};
            \node at (0.583, -0.2) {Flow Priors};
            \node at (0.750, -0.2) {DPIR};
            \node at (0.916, -0.2) {RED};
        \end{scope}
    \end{tikzpicture}
    \caption{Qualitative comparison of image restoration results on AFHQ-Cat (random inpainting).}
    \label{fig:quality_comparisons_app_afhq_random_inpainting}
\end{figure}

\begin{figure}[htbp]
    \centering
    \begin{tikzpicture}
        \node[anchor=south west, inner sep=0] (row1) at (0,7.7) {\includegraphics[width=\linewidth]{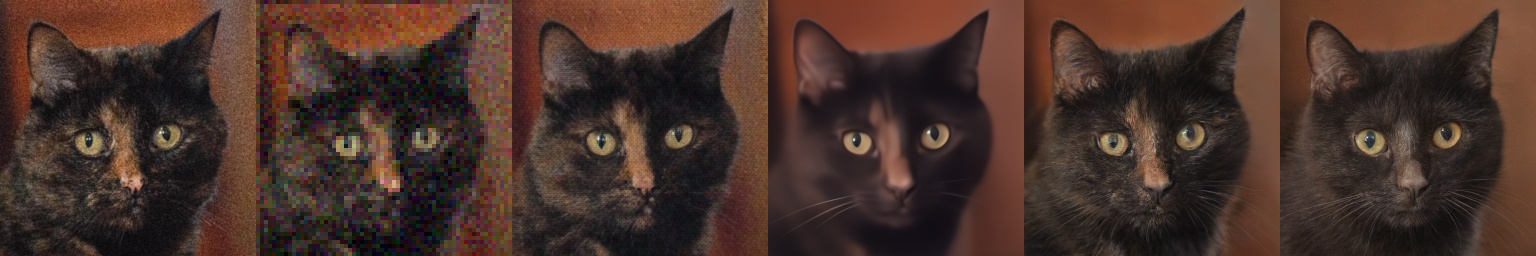}};
        \begin{scope}[shift={(row1.north west)}, x=\linewidth]
            \tikzset{every node/.style={anchor=base, font=\scriptsize, text height=1.5ex, text depth=.25ex}}
            \node at (0.083, 0.05) {Clean};
            \node at (0.250, 0.05) {Degraded};
            \node at (0.416, 0.05) {\coolname{}};
            \node at (0.583, 0.05) {DDRM};
            \node at (0.750, 0.05) {DPS};
            \node at (0.916, 0.05) {PnP-Diff};
        \end{scope}

        \node[anchor=south west, inner sep=0] (row2) at (0,5.37) {\includegraphics[width=\linewidth]{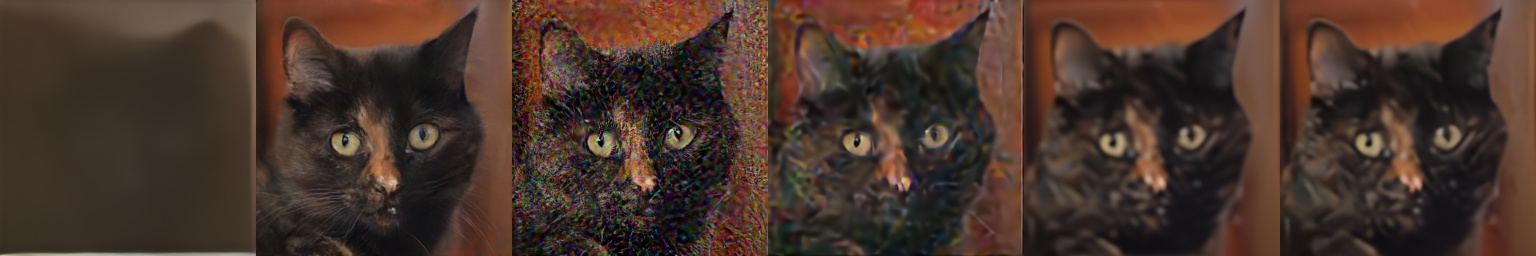}};
        \begin{scope}[shift={(row2.south west)}, x=\linewidth]
            \tikzset{every node/.style={anchor=base, font=\scriptsize, text height=1.5ex, text depth=.25ex}}
            \node at (0.083, -0.2) {PnP-Flow};
            \node at (0.250, -0.2) {OT-ODE};
            \node at (0.416, -0.2) {D-Flow};
            \node at (0.583, -0.2) {Flow Priors};
            \node at (0.750, -0.2) {DPIR};
            \node at (0.916, -0.2) {RED};
        \end{scope}

        \node[anchor=south west, inner sep=0] (row3) at (0,2.33) {\includegraphics[width=\linewidth]{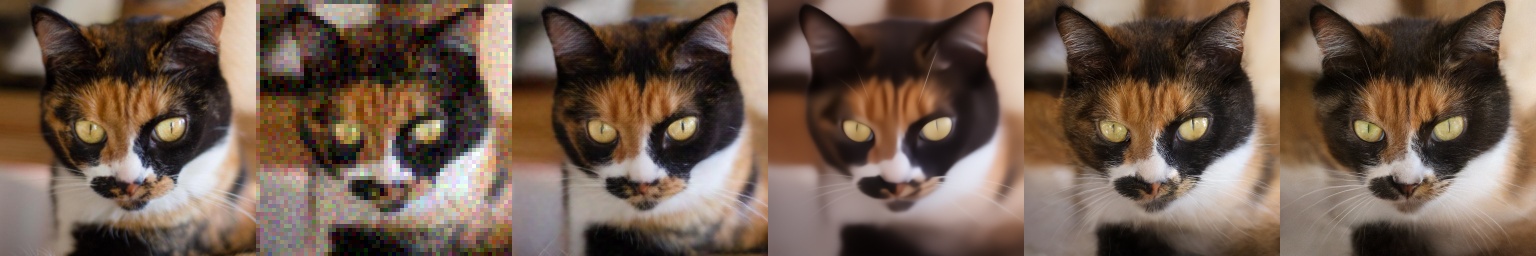}};
        \begin{scope}[shift={(row3.north west)}, x=\linewidth]
            \tikzset{every node/.style={anchor=base, font=\scriptsize, text height=1.5ex, text depth=.25ex}}
            \node at (0.083, 0.05) {Clean};
            \node at (0.250, 0.05) {Degraded};
            \node at (0.416, 0.05) {\coolname{}};
            \node at (0.583, 0.05) {DDRM};
            \node at (0.750, 0.05) {DPS};
            \node at (0.916, 0.05) {PnP-Diff};
        \end{scope}

        \node[anchor=south west, inner sep=0] (row4) at (0,0) {\includegraphics[width=\linewidth]{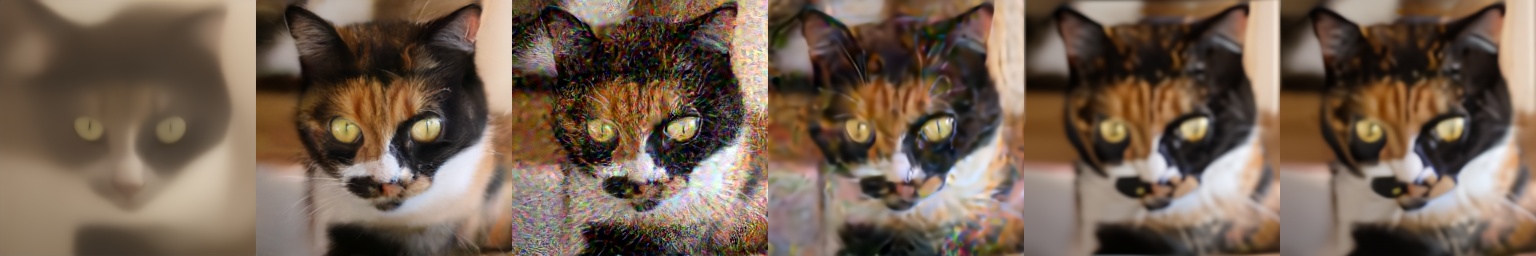}};
        \begin{scope}[shift={(row4.south west)}, x=\linewidth]
            \tikzset{every node/.style={anchor=base, font=\scriptsize, text height=1.5ex, text depth=.25ex}}
            \node at (0.083, -0.2) {PnP-Flow};
            \node at (0.250, -0.2) {OT-ODE};
            \node at (0.416, -0.2) {D-Flow};
            \node at (0.583, -0.2) {Flow Priors};
            \node at (0.750, -0.2) {DPIR};
            \node at (0.916, -0.2) {RED};
        \end{scope}
    \end{tikzpicture}
    \caption{Qualitative comparison of image restoration results on AFHQ-Cat (super-resolution).}
    \label{fig:quality_comparisons_app_afhq_superresolution}
\end{figure}

\begin{figure}[t]
    \centering
    \begin{tikzpicture}
        \node[anchor=south west, inner sep=0] (row1) at (0,6.9) {\includegraphics[width=\linewidth]{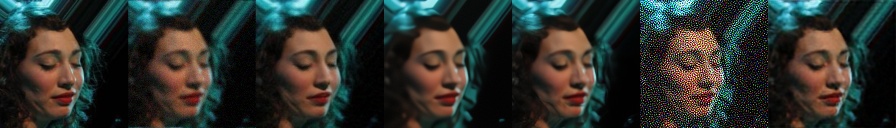}};
        \begin{scope}[shift={(row1.north west)}, x=\linewidth]
            \tikzset{every node/.style={anchor=base, font=\tiny, text height=1.5ex, text depth=.25ex}}
            \node at (0.071, 0.05) {Clean};
            \node at (0.214, 0.05) {Degraded};
            \node at (0.357, 0.05) {\coolname{}};
            \node at (0.500, 0.05) {DDRM};
            \node at (0.643, 0.05) {DPS};
            \node at (0.786, 0.05) {PnP-Diff};
            \node at (0.928, 0.05) {PnP-GS};
        \end{scope}

        \node[anchor=south west, inner sep=0] (row2) at (0,4.9) {\includegraphics[width=0.857\linewidth]{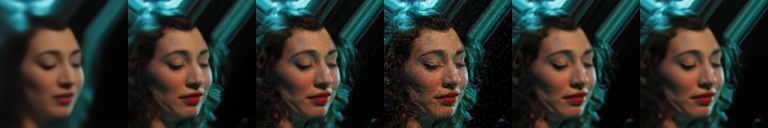}};
        \begin{scope}[shift={(row2.south west)}, x=\linewidth]
            \tikzset{every node/.style={anchor=north, font=\tiny, text height=1.5ex}}
            \node at (0.071, 0.15) {PnP-Flow};
            \node at (0.214, 0.15) {OT-ODE};
            \node at (0.357, 0.15) {D-Flow};
            \node at (0.500, 0.15) {Flow Priors};
            \node at (0.643, 0.15) {DPIR};
            \node at (0.786, 0.15) {RED};
        \end{scope}

        \node[anchor=south west, inner sep=0] (row3) at (0,2.3) {\includegraphics[width=\linewidth]{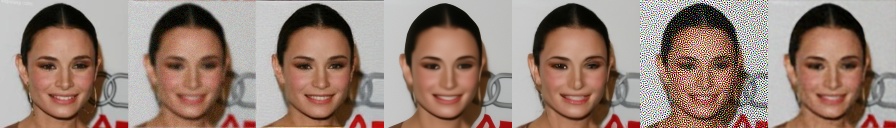}};
        \begin{scope}[shift={(row3.north west)}, x=\linewidth]
            \tikzset{every node/.style={anchor=base, font=\tiny, text height=1.5ex, text depth=.25ex}}
            \node at (0.071, 0.05) {Clean};
            \node at (0.214, 0.05) {Degraded};
            \node at (0.357, 0.05) {\coolname{}};
            \node at (0.500, 0.05) {DDRM};
            \node at (0.643, 0.05) {DPS};
            \node at (0.786, 0.05) {PnP-Diff};
            \node at (0.928, 0.05) {PnP-GS};
        \end{scope}

        \node[anchor=south west, inner sep=0] (row4) at (0,0.3) {\includegraphics[width=0.857\linewidth]{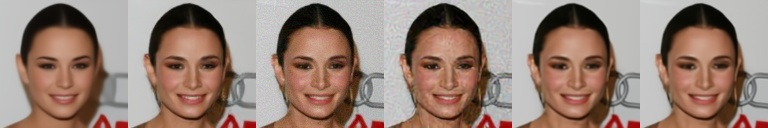}};
        \begin{scope}[shift={(row4.south west)}, x=\linewidth]
            \tikzset{every node/.style={anchor=north, font=\tiny, text height=1.5ex}}
            \node at (0.071, 0.15) {PnP-Flow};
            \node at (0.214, 0.15) {OT-ODE};
            \node at (0.357, 0.15) {D-Flow};
            \node at (0.500, 0.15) {Flow Priors};
            \node at (0.643, 0.15) {DPIR};
            \node at (0.786, 0.15) {RED};
        \end{scope}
    \end{tikzpicture}
    \caption{Qualitative comparison of image restoration results on CelebA (deblurring).}
    \label{fig:quality_comparisons_app_celeba_deblurring}
\end{figure}

\begin{figure}[t]
    \centering
    \begin{tikzpicture}
        \node[anchor=south west, inner sep=0] (row1) at (0,6.9) {\includegraphics[width=\linewidth]{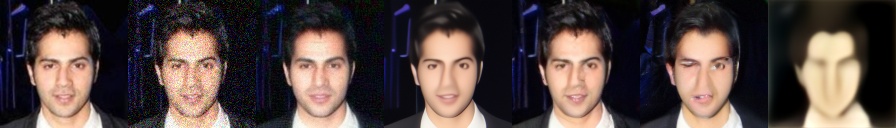}};
        \begin{scope}[shift={(row1.north west)}, x=\linewidth]
            \tikzset{every node/.style={anchor=base, font=\tiny, text height=1.5ex, text depth=.25ex}}
            \node at (0.071, 0.05) {Clean};
            \node at (0.214, 0.05) {Degraded};
            \node at (0.357, 0.05) {\coolname{}};
            \node at (0.500, 0.05) {DDRM};
            \node at (0.643, 0.05) {DPS};
            \node at (0.786, 0.05) {PnP-Diff};
            \node at (0.928, 0.05) {PnP-GS};
        \end{scope}

        \node[anchor=south west, inner sep=0] (row2) at (0,4.9) {\includegraphics[width=0.857\linewidth]{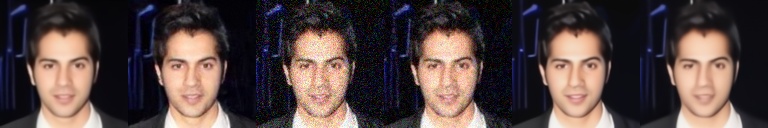}};
        \begin{scope}[shift={(row2.south west)}, x=\linewidth]
            \tikzset{every node/.style={anchor=north, font=\tiny, text height=1.5ex}}
            \node at (0.071, 0.15) {PnP-Flow};
            \node at (0.214, 0.15) {OT-ODE};
            \node at (0.357, 0.15) {D-Flow};
            \node at (0.500, 0.15) {Flow Priors};
            \node at (0.643, 0.15) {DPIR};
            \node at (0.786, 0.15) {RED};
        \end{scope}

        \node[anchor=south west, inner sep=0] (row3) at (0,2.3) {\includegraphics[width=\linewidth]{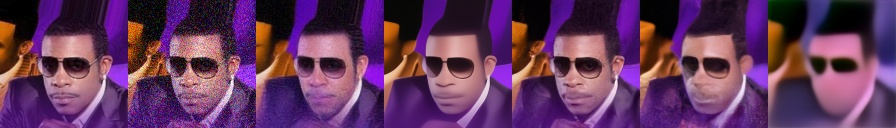}};
        \begin{scope}[shift={(row3.north west)}, x=\linewidth]
            \tikzset{every node/.style={anchor=base, font=\tiny, text height=1.5ex, text depth=.25ex}}
            \node at (0.071, 0.05) {Clean};
            \node at (0.214, 0.05) {Degraded};
            \node at (0.357, 0.05) {\coolname{}};
            \node at (0.500, 0.05) {DDRM};
            \node at (0.643, 0.05) {DPS};
            \node at (0.786, 0.05) {PnP-Diff};
            \node at (0.928, 0.05) {PnP-GS};
        \end{scope}

        \node[anchor=south west, inner sep=0] (row4) at (0,0.3) {\includegraphics[width=0.857\linewidth]{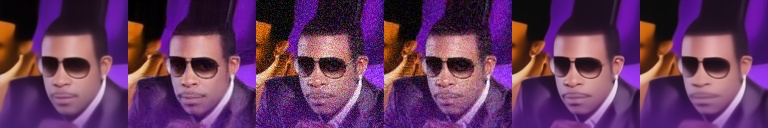}};
        \begin{scope}[shift={(row4.south west)}, x=\linewidth]
            \tikzset{every node/.style={anchor=north, font=\tiny, text height=1.5ex}}
            \node at (0.071, 0.15) {PnP-Flow};
            \node at (0.214, 0.15) {OT-ODE};
            \node at (0.357, 0.15) {D-Flow};
            \node at (0.500, 0.15) {Flow Priors};
            \node at (0.643, 0.15) {DPIR};
            \node at (0.786, 0.15) {RED};
        \end{scope}
    \end{tikzpicture}
    \caption{Qualitative comparison of image restoration results on CelebA (denoising).}
    \label{fig:quality_comparisons_app_celeba_denoising}
\end{figure}

\begin{figure}[t]
    \centering
    \begin{tikzpicture}
        \node[anchor=south west, inner sep=0] (row1) at (0,6.9) {\includegraphics[width=\linewidth]{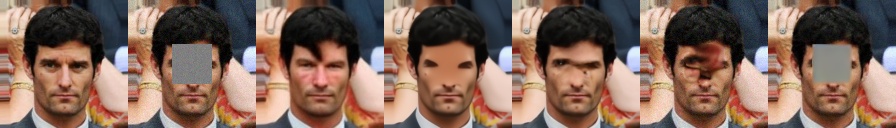}};
        \begin{scope}[shift={(row1.north west)}, x=\linewidth]
            \tikzset{every node/.style={anchor=base, font=\tiny, text height=1.5ex, text depth=.25ex}}
            \node at (0.071, 0.05) {Clean};
            \node at (0.214, 0.05) {Degraded};
            \node at (0.357, 0.05) {\coolname{}};
            \node at (0.500, 0.05) {DDRM};
            \node at (0.643, 0.05) {DPS};
            \node at (0.786, 0.05) {PnP-Diff};
            \node at (0.928, 0.05) {PnP-GS};
        \end{scope}

        \node[anchor=south west, inner sep=0] (row2) at (0,4.9) {\includegraphics[width=0.857\linewidth]{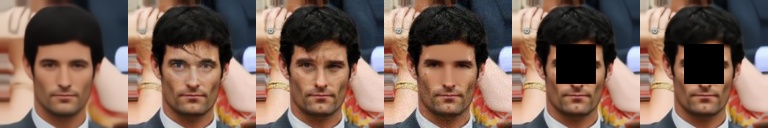}};
        \begin{scope}[shift={(row2.south west)}, x=\linewidth]
            \tikzset{every node/.style={anchor=north, font=\tiny, text height=1.5ex}}
            \node at (0.071, 0.15) {PnP-Flow};
            \node at (0.214, 0.15) {OT-ODE};
            \node at (0.357, 0.15) {D-Flow};
            \node at (0.500, 0.15) {Flow Priors};
            \node at (0.643, 0.15) {DPIR};
            \node at (0.786, 0.15) {RED};
        \end{scope}

        \node[anchor=south west, inner sep=0] (row3) at (0,2.3) {\includegraphics[width=\linewidth]{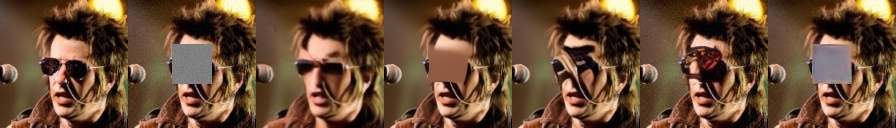}};
        \begin{scope}[shift={(row3.north west)}, x=\linewidth]
            \tikzset{every node/.style={anchor=base, font=\tiny, text height=1.5ex, text depth=.25ex}}
            \node at (0.071, 0.05) {Clean};
            \node at (0.214, 0.05) {Degraded};
            \node at (0.357, 0.05) {\coolname{}};
            \node at (0.500, 0.05) {DDRM};
            \node at (0.643, 0.05) {DPS};
            \node at (0.786, 0.05) {PnP-Diff};
            \node at (0.928, 0.05) {PnP-GS};
        \end{scope}

        \node[anchor=south west, inner sep=0] (row4) at (0,0.3) {\includegraphics[width=0.857\linewidth]{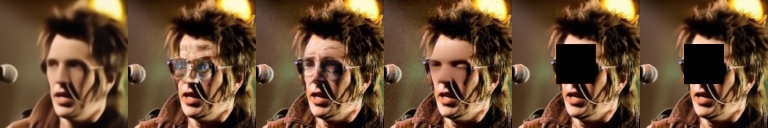}};
        \begin{scope}[shift={(row4.south west)}, x=\linewidth]
            \tikzset{every node/.style={anchor=north, font=\tiny, text height=1.5ex}}
            \node at (0.071, 0.15) {PnP-Flow};
            \node at (0.214, 0.15) {OT-ODE};
            \node at (0.357, 0.15) {D-Flow};
            \node at (0.500, 0.15) {Flow Priors};
            \node at (0.643, 0.15) {DPIR};
            \node at (0.786, 0.15) {RED};
        \end{scope}
    \end{tikzpicture}
    \caption{Qualitative comparison of image restoration results on CelebA (inpainting).}
    \label{fig:quality_comparisons_app_celeba_inpainting}
\end{figure}

\begin{figure}[t]
    \centering
    \begin{tikzpicture}
        \node[anchor=south west, inner sep=0] (row1) at (0,6.9) {\includegraphics[width=\linewidth]{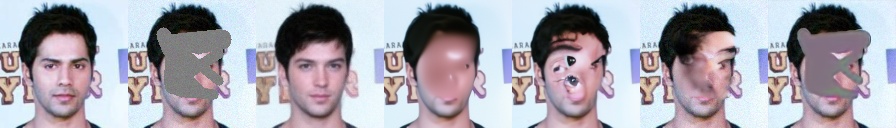}};
        \begin{scope}[shift={(row1.north west)}, x=\linewidth]
            \tikzset{every node/.style={anchor=base, font=\tiny, text height=1.5ex, text depth=.25ex}}
            \node at (0.071, 0.05) {Clean};
            \node at (0.214, 0.05) {Degraded};
            \node at (0.357, 0.05) {\coolname{}};
            \node at (0.500, 0.05) {DDRM};
            \node at (0.643, 0.05) {DPS};
            \node at (0.786, 0.05) {PnP-Diff};
            \node at (0.928, 0.05) {PnP-GS};
        \end{scope}

        \node[anchor=south west, inner sep=0] (row2) at (0,4.9) {\includegraphics[width=0.857\linewidth]{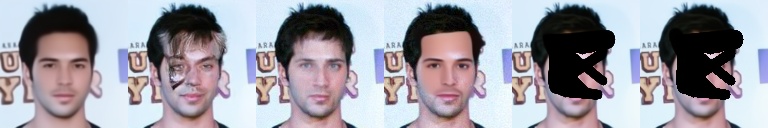}};
        \begin{scope}[shift={(row2.south west)}, x=\linewidth]
            \tikzset{every node/.style={anchor=north, font=\tiny, text height=1.5ex}}
            \node at (0.071, 0.15) {PnP-Flow};
            \node at (0.214, 0.15) {OT-ODE};
            \node at (0.357, 0.15) {D-Flow};
            \node at (0.500, 0.15) {Flow Priors};
            \node at (0.643, 0.15) {DPIR};
            \node at (0.786, 0.15) {RED};
        \end{scope}

        \node[anchor=south west, inner sep=0] (row3) at (0,2.3) {\includegraphics[width=\linewidth]{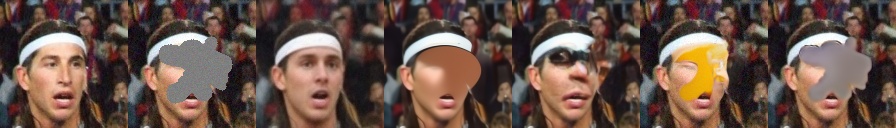}};
        \begin{scope}[shift={(row3.north west)}, x=\linewidth]
            \tikzset{every node/.style={anchor=base, font=\tiny, text height=1.5ex, text depth=.25ex}}
            \node at (0.071, 0.05) {Clean};
            \node at (0.214, 0.05) {Degraded};
            \node at (0.357, 0.05) {\coolname{}};
            \node at (0.500, 0.05) {DDRM};
            \node at (0.643, 0.05) {DPS};
            \node at (0.786, 0.05) {PnP-Diff};
            \node at (0.928, 0.05) {PnP-GS};
        \end{scope}

        \node[anchor=south west, inner sep=0] (row4) at (0,0.3) {\includegraphics[width=0.857\linewidth]{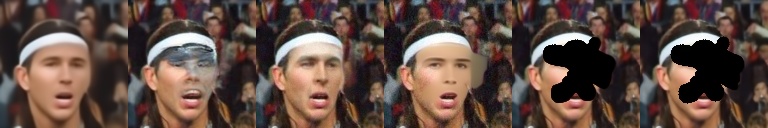}};
        \begin{scope}[shift={(row4.south west)}, x=\linewidth]
            \tikzset{every node/.style={anchor=north, font=\tiny, text height=1.5ex}}
            \node at (0.071, 0.15) {PnP-Flow};
            \node at (0.214, 0.15) {OT-ODE};
            \node at (0.357, 0.15) {D-Flow};
            \node at (0.500, 0.15) {Flow Priors};
            \node at (0.643, 0.15) {DPIR};
            \node at (0.786, 0.15) {RED};
        \end{scope}
    \end{tikzpicture}
    \caption{Qualitative comparison of image restoration results on CelebA (paintbrush-inpainting).}
    \label{fig:quality_comparisons_app_celeba_paintbrush_inpainting}
\end{figure}

\begin{figure}[t]
    \centering
    \begin{tikzpicture}
        \node[anchor=south west, inner sep=0] (row1) at (0,6.9) {\includegraphics[width=\linewidth]{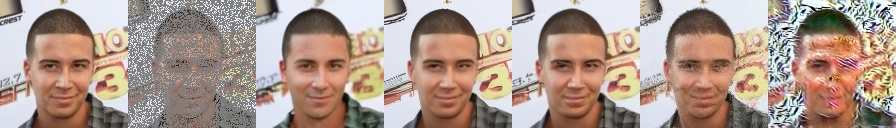}};
        \begin{scope}[shift={(row1.north west)}, x=\linewidth]
            \tikzset{every node/.style={anchor=base, font=\tiny, text height=1.5ex, text depth=.25ex}}
            \node at (0.071, 0.05) {Clean};
            \node at (0.214, 0.05) {Degraded};
            \node at (0.357, 0.05) {\coolname{}};
            \node at (0.500, 0.05) {DDRM};
            \node at (0.643, 0.05) {DPS};
            \node at (0.786, 0.05) {PnP-Diff};
            \node at (0.928, 0.05) {PnP-GS};
        \end{scope}

        \node[anchor=south west, inner sep=0] (row2) at (0,4.9) {\includegraphics[width=0.857\linewidth]{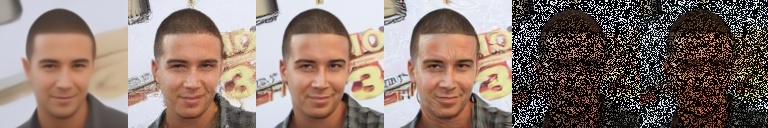}};
        \begin{scope}[shift={(row2.south west)}, x=\linewidth]
            \tikzset{every node/.style={anchor=north, font=\tiny, text height=1.5ex}}
            \node at (0.071, 0.15) {PnP-Flow};
            \node at (0.214, 0.15) {OT-ODE};
            \node at (0.357, 0.15) {D-Flow};
            \node at (0.500, 0.15) {Flow Priors};
            \node at (0.643, 0.15) {DPIR};
            \node at (0.786, 0.15) {RED};
        \end{scope}

        \node[anchor=south west, inner sep=0] (row3) at (0,2.3) {\includegraphics[width=\linewidth]{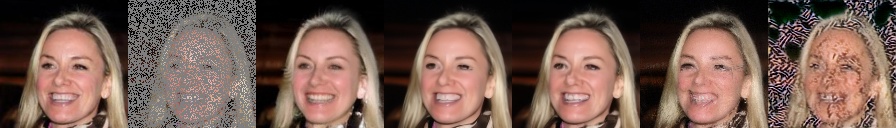}};
        \begin{scope}[shift={(row3.north west)}, x=\linewidth]
            \tikzset{every node/.style={anchor=base, font=\tiny, text height=1.5ex, text depth=.25ex}}
            \node at (0.071, 0.05) {Clean};
            \node at (0.214, 0.05) {Degraded};
            \node at (0.357, 0.05) {\coolname{}};
            \node at (0.500, 0.05) {DDRM};
            \node at (0.643, 0.05) {DPS};
            \node at (0.786, 0.05) {PnP-Diff};
            \node at (0.928, 0.05) {PnP-GS};
        \end{scope}

        \node[anchor=south west, inner sep=0] (row4) at (0,0.3) {\includegraphics[width=0.857\linewidth]{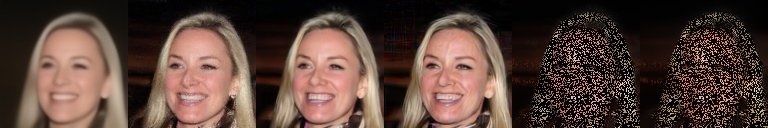}};
        \begin{scope}[shift={(row4.south west)}, x=\linewidth]
            \tikzset{every node/.style={anchor=north, font=\tiny, text height=1.5ex}}
            \node at (0.071, 0.15) {PnP-Flow};
            \node at (0.214, 0.15) {OT-ODE};
            \node at (0.357, 0.15) {D-Flow};
            \node at (0.500, 0.15) {Flow Priors};
            \node at (0.643, 0.15) {DPIR};
            \node at (0.786, 0.15) {RED};
        \end{scope}
    \end{tikzpicture}
    \caption{Qualitative comparison of image restoration results on CelebA (random-inpainting).}
    \label{fig:quality_comparisons_app_celeba_random_inpainting}
\end{figure}

\begin{figure}[t]
    \centering
    \begin{tikzpicture}
        \node[anchor=south west, inner sep=0] (row1) at (0,6.9) {\includegraphics[width=\linewidth]{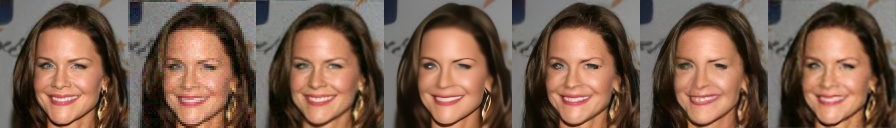}};
        \begin{scope}[shift={(row1.north west)}, x=\linewidth]
            \tikzset{every node/.style={anchor=base, font=\tiny, text height=1.5ex, text depth=.25ex}}
            \node at (0.071, 0.05) {Clean};
            \node at (0.214, 0.05) {Degraded};
            \node at (0.357, 0.05) {\coolname{}};
            \node at (0.500, 0.05) {DDRM};
            \node at (0.643, 0.05) {DPS};
            \node at (0.786, 0.05) {PnP-Diff};
            \node at (0.928, 0.05) {PnP-GS};
        \end{scope}

        \node[anchor=south west, inner sep=0] (row2) at (0,4.9) {\includegraphics[width=0.857\linewidth]{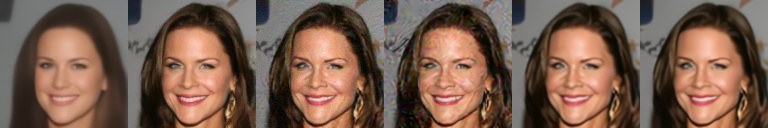}};
        \begin{scope}[shift={(row2.south west)}, x=\linewidth]
            \tikzset{every node/.style={anchor=north, font=\tiny, text height=1.5ex}}
            \node at (0.071, 0.15) {PnP-Flow};
            \node at (0.214, 0.15) {OT-ODE};
            \node at (0.357, 0.15) {D-Flow};
            \node at (0.500, 0.15) {Flow Priors};
            \node at (0.643, 0.15) {DPIR};
            \node at (0.786, 0.15) {RED};
        \end{scope}

        \node[anchor=south west, inner sep=0] (row3) at (0,2.3) {\includegraphics[width=\linewidth]{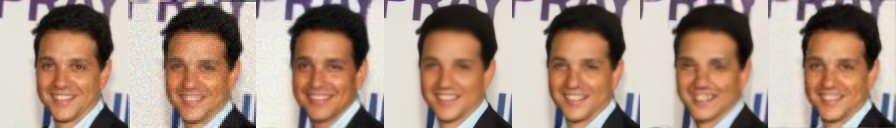}};
        \begin{scope}[shift={(row3.north west)}, x=\linewidth]
            \tikzset{every node/.style={anchor=base, font=\tiny, text height=1.5ex, text depth=.25ex}}
            \node at (0.071, 0.05) {Clean};
            \node at (0.214, 0.05) {Degraded};
            \node at (0.357, 0.05) {\coolname{}};
            \node at (0.500, 0.05) {DDRM};
            \node at (0.643, 0.05) {DPS};
            \node at (0.786, 0.05) {PnP-Diff};
            \node at (0.928, 0.05) {PnP-GS};
        \end{scope}

        \node[anchor=south west, inner sep=0] (row4) at (0,0.3) {\includegraphics[width=0.857\linewidth]{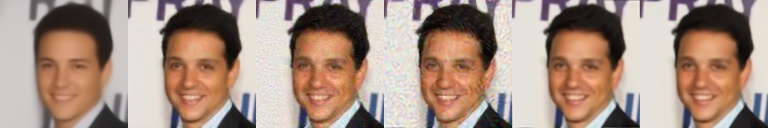}};
        \begin{scope}[shift={(row4.south west)}, x=\linewidth]
            \tikzset{every node/.style={anchor=north, font=\tiny, text height=1.5ex}}
            \node at (0.071, 0.15) {PnP-Flow};
            \node at (0.214, 0.15) {OT-ODE};
            \node at (0.357, 0.15) {D-Flow};
            \node at (0.500, 0.15) {Flow Priors};
            \node at (0.643, 0.15) {DPIR};
            \node at (0.786, 0.15) {RED};
        \end{scope}
    \end{tikzpicture}
    \caption{Qualitative comparison of image restoration results on CelebA (super-resolution).}
    \label{fig:quality_comparisons_app_celeba_superresolution}
\end{figure}

Figures \ref{fig:quality_comparisons_app_afhq_deblurring}-\ref{fig:quality_comparisons_app_afhq_superresolution} 
and Figures \ref{fig:quality_comparisons_app_celeba_deblurring}-\ref{fig:quality_comparisons_app_celeba_superresolution}
illustrate qualitative comparisons of all methods on AFHQ-Cat and CelebA, respectively.

\Cref{tab:res_app_afhq_1,tab:res_app_afhq_2}, and \Cref{tab:res_app_celeba_1,tab:res_app_celeba_2} provide PSNR and LPIPS for distortion metrics, and KID for perception metric, for all methods AFHQ-Cat and CelebA, respectively.

\cref{tab:timing_by_method} provides running time (in seconds) of all methods for a single image. 

\subsection{Qualitative comparisons}

\clearpage

%% file: sections/C_tables.tex
\subsection{Quantitative results}
\label{app:results}

\begin{table*}[h]
\centering
\resizebox{\textwidth}{!}{%
\begin{tabular}{lccccccccc}
\toprule
\textbf{Method} & \multicolumn{3}{c}{\textbf{Denoising}} & \multicolumn{3}{c}{\textbf{Deblurring}} & \multicolumn{3}{c}{\textbf{Inpainting}} \\
\cmidrule(lr){2-4} \cmidrule(lr){5-7} \cmidrule(lr){8-10}
 & \textbf{KID $\downarrow$} & \textbf{LPIPS $\downarrow$} & \textbf{PSNR $\uparrow$} & \textbf{KID $\downarrow$} & \textbf{LPIPS $\downarrow$} & \textbf{PSNR $\uparrow$} & \textbf{KID $\downarrow$} & \textbf{LPIPS $\downarrow$} & \textbf{PSNR $\uparrow$} \\
\midrule
DPS & 3.63 & 0.159 & 26.38 & 2.69 & 0.178 & 24.72 & 3.91 & 0.107 & 25.85 \\
OT-ODE & 7.71 & 0.146 & 27.32 & 6.74 & 0.155 & 26.79 & 8.53 & 0.129 & 22.95 \\
DiffPIR & 5.74 & 0.235 & 23.05 & 128.68 & 0.584 & 16.71 & 15.06 & 0.267 & 23.72 \\
PnP-Flow & 36.74 & 0.259 & 29.49 & 64.01 & 0.446 & 25.91 & 58.89 & 0.328 & 25.57 \\
DDRM & 48.83 & 0.336 & 26.74 & 41.93 & 0.325 & 27.34 & 40.21 & 0.220 & 26.54 \\
RED & 55.42 & 0.414 & 25.99 & 49.44 & 0.398 & 28.25 & 106.65 & 0.206 & 17.20 \\
DPIR & 66.70 & 0.336 & 27.97 & 40.84 & 0.401 & 28.08 & 122.08 & 0.230 & 17.27 \\
Flow Priors & 38.86 & 0.700 & 19.33 & 21.18 & 0.297 & 26.46 & 30.95 & 0.301 & 24.34 \\
D-Flow & 60.17 & 1.039 & 14.97 & 7.81 & 0.170 & 27.20 & 20.70 & 0.224 & 23.71 \\
\coolname & 4.09 & 0.174 & 22.84 & 2.56 & 0.118 & 25.75 & 6.68 & 0.103 & 25.12 \\
\bottomrule
\end{tabular}%
}
\caption{Quantitative evaluation on AFHQ cats. KID is scaled by $\times 10^3$.}
\label{tab:res_app_afhq_1}
\end{table*}

\begin{table*}[h]
\centering
\resizebox{\textwidth}{!}{%
\begin{tabular}{lccccccccc}
\toprule
\textbf{Method} & \multicolumn{3}{c}{\textbf{Paintbrush Inp.}} & \multicolumn{3}{c}{\textbf{Random Inp.}} & \multicolumn{3}{c}{\textbf{Superresolution}} \\
\cmidrule(lr){2-4} \cmidrule(lr){5-7} \cmidrule(lr){8-10}
 & \textbf{KID $\downarrow$} & \textbf{LPIPS $\downarrow$} & \textbf{PSNR $\uparrow$} & \textbf{KID $\downarrow$} & \textbf{LPIPS $\downarrow$} & \textbf{PSNR $\uparrow$} & \textbf{KID $\downarrow$} & \textbf{LPIPS $\downarrow$} & \textbf{PSNR $\uparrow$} \\
\midrule
DPS & 4.53 & 0.103 & 26.92 & 3.15 & 0.100 & 29.26 & 3.50 & 0.201 & 24.32 \\
OT-ODE & 9.10 & 0.120 & 24.14 & 4.34 & 0.118 & 28.80 & 6.40 & 0.173 & 26.34 \\
DiffPIR & 16.69 & 0.263 & 24.26 & 9.55 & 0.159 & 27.26 & 5.59 & 0.242 & 22.60 \\
PnP-Flow & 57.43 & 0.321 & 26.34 & 66.44 & 0.438 & 25.16 & 98.44 & 0.666 & 18.09 \\
DDRM & 40.20 & 0.213 & 27.55 & 20.93 & 0.114 & 32.32 & 50.64 & 0.371 & 25.42 \\
RED & 59.57 & 0.159 & 18.19 & 360.24 & 1.171 & 8.31 & 45.91 & 0.373 & 27.95 \\
DPIR & 96.43 & 0.198 & 18.29 & 333.30 & 1.139 & 7.97 & 47.11 & 0.398 & 27.42 \\
Flow Priors & 23.60 & 0.278 & 25.23 & 16.51 & 0.145 & 29.77 & 55.70 & 0.417 & 24.41 \\
D-Flow & 19.14 & 0.217 & 24.40 & 6.64 & 0.061 & 32.24 & 35.15 & 0.568 & 21.28 \\
\coolname & 6.91 & 0.099 & 25.73 & 3.71 & 0.085 & 27.62 & 3.47 & 0.151 & 24.24 \\
\bottomrule
\end{tabular}%
}
\caption{Quantitative evaluation on AFHQ cats. KID is scaled by $\times 10^3$.}
\label{tab:res_app_afhq_2}
\end{table*}

\clearpage

\begin{table*}[h]
\centering
\resizebox{\textwidth}{!}{%
\begin{tabular}{lccccccccc}
\toprule
\textbf{Method} & \multicolumn{3}{c}{\textbf{Denoising}} & \multicolumn{3}{c}{\textbf{Deblurring}} & \multicolumn{3}{c}{\textbf{Inpainting}} \\
\cmidrule(lr){2-4} \cmidrule(lr){5-7} \cmidrule(lr){8-10}
 & \textbf{KID $\downarrow$} & \textbf{LPIPS $\downarrow$} & \textbf{PSNR $\uparrow$} & \textbf{KID $\downarrow$} & \textbf{LPIPS $\downarrow$} & \textbf{PSNR $\uparrow$} & \textbf{KID $\downarrow$} & \textbf{LPIPS $\downarrow$} & \textbf{PSNR $\uparrow$} \\
\midrule
OT-ODE & 7.70 & 0.044 & 30.47 & 5.90 & 0.037 & 32.80 & 8.29 & 0.050 & 27.57 \\
DPS & 17.52 & 0.041 & 31.30 & 14.67 & 0.041 & 32.09 & 16.16 & 0.055 & 25.14 \\
DPIR & 44.45 & 0.071 & 32.02 & 19.48 & 0.046 & 35.42 & 121.67 & 0.250 & 14.88 \\
PnP-Flow & 45.58 & 0.114 & 29.95 & 57.95 & 0.184 & 27.90 & 52.23 & 0.136 & 27.61 \\
DiffPIR & 64.02 & 0.109 & 26.47 & 272.39 & 0.951 & 10.73 & 74.04 & 0.105 & 22.61 \\
RED & 62.99 & 0.126 & 30.02 & 29.82 & 0.049 & 34.81 & 136.61 & 0.261 & 14.85 \\
Flow Priors & 86.68 & 0.234 & 24.73 & 22.96 & 0.064 & 31.52 & 12.27 & 0.058 & 29.65 \\
DDRM & 103.42 & 0.113 & 29.89 & 65.43 & 0.067 & 32.95 & 68.42 & 0.089 & 26.80 \\
D-Flow & 95.72 & 0.390 & 20.72 & 67.70 & 0.027 & 30.60 & 48.90 & 0.025 & 30.74 \\
PnP-GS & 187.39 & 0.418 & 21.46 & 21.22 & 0.057 & 33.41 & 128.87 & 0.155 & 22.97 \\
\coolname & 57.10 & 0.147 & 25.45 & 7.95 & 0.057 & 31.13 & 34.38 & 0.0522 & 27.23 \\
\bottomrule
\end{tabular}%
}
\caption{Quantitative evaluation on CelebA. KID is scaled by $\times 10^3$.}
\label{tab:res_app_celeba_1}
\end{table*}

\begin{table*}[h]
\centering
\resizebox{\textwidth}{!}{%
\begin{tabular}{lrrrrrrrrr}
\toprule
\textbf{Method} & \multicolumn{3}{c}{\textbf{Paintbrush Inp.}} & \multicolumn{3}{c}{\textbf{Random Inp.}} & \multicolumn{3}{c}{\textbf{Superresolution}} \\
\cmidrule(lr){2-4} \cmidrule(lr){5-7} \cmidrule(lr){8-10}
 & \textbf{KID $\downarrow$} & \textbf{LPIPS $\downarrow$} & \textbf{PSNR $\uparrow$} & \textbf{KID $\downarrow$} & \textbf{LPIPS $\downarrow$} & \textbf{PSNR $\uparrow$} & \textbf{KID $\downarrow$} & \textbf{LPIPS $\downarrow$} & \textbf{PSNR $\uparrow$} \\
\midrule
OT-ODE & 12.01 & 0.099 & 22.93 & 20.53 & 0.084 & 27.61 & 6.28 & 0.047 & 31.30 \\
DPS & 22.72 & 0.104 & 21.49 & 7.79 & 0.026 & 31.91 & 15.86 & 0.045 & 31.63 \\
DPIR & 162.28 & 0.291 & 12.54 & 425.53 & 1.018 & 8.45 & 25.84 & 0.067 & 32.75 \\
PnP-Flow & 54.00 & 0.160 & 25.06 & 73.93 & 0.233 & 23.32 & 80.71 & 0.272 & 22.14 \\
PnP-Diff & 83.56 & 0.175 & 19.07 & 81.34 & 0.118 & 26.70 & 49.54 & 0.067 & 28.76 \\
RED & 224.26 & 0.304 & 12.47 & 434.20 & 1.015 & 8.87 & 29.05 & 0.062 & 33.65 \\
Flow Priors & 33.91 & 0.097 & 24.97 & 18.45 & 0.072 & 29.05 & 33.75 & 0.119 & 28.53 \\
DDRM & 86.61 & 0.160 & 23.21 & 26.63 & 0.041 & 31.04 & 87.84 & 0.091 & 30.51 \\
D-Flow & 51.96 & 0.045 & 27.43 & 7.08 & 0.021 & 33.28 & 71.54 & 0.044 & 28.62 \\
PnP-GS & 112.76 & 0.233 & 19.45 & 343.59 & 0.638 & 12.46 & 24.44 & 0.072 & 31.33 \\
\coolname & 50.00 & 0.088 & 23.48 & 22.39 & 0.080 & 26.06 & 13.80 & 0.067 & 29.46 \\
\bottomrule
\end{tabular}%
}
\caption{Quantitative evaluation on CelebA. KID is scaled by $\times 10^3$.}
\label{tab:res_app_celeba_2}
\end{table*}

\clearpage

\begin{table*}[h]
\centering
\begin{tabular}{lc}
\toprule
\textbf{Method} & \textbf{Runtime [sec]} \\
\midrule
OT-ODE & 2.29 \\
DPS & 17.77 \\
DPIR & 0.044 \\
PnP-Flow & 3.20 \\
DiffPIR & 0.83 \\
RED & 0.37 \\
Flow Priors & 29.88 \\
DDRM & 0.81 \\
D-Flow & 183.18 \\
PnP-GS & 0.31 \\
\coolname (1 step) & 0.016 \\
\coolname (3 step) & 0.043 \\
\coolname (5 step) & 0.073 \\
\coolname (10 step) & 0.14 \\
\coolname (20 step) & 0.28 \\
\bottomrule
\end{tabular}
\caption{\textbf{Average inference time per image.} Reported per-image wall-clock runtimes were measured using a single NVIDIA L40S GPU for all methods.}
\label{tab:timing_by_method}
\end{table*}

%% file: sections/D_implementation.tex
\section{Implementation details}
\label{app:implementation}

\coolname (\cref{alg:encdec}) is primarily governed by two key hyper-parameters: the data prox regularization coefficient, $\lambda$, and the injected latent noise scale, $\sigma$. Note that $\sigma$ denotes the relative noise strength with respect to the maximum noise level the prior was trained on (equivalent to $r$ in \cite{yue2026image}). In addition to these parameters, the number of forward steps can be adjusted; however, this serves as a flexible parameter rather than a strict constraint due to the anytime restoration capabilities of \coolname.

Furthermore, the algorithm requires an initialization strategy for the starting state, $x_0$, which we tailor to the specific degradation. For denoising and deblurring, we apply the adjoint operator. For inpainting problems, we utilize a GPU-friendly, convolution-based median filter approximation. Finally, for super-resolution, we initialize using standard bicubic upsampling.

The optimal values for $\lambda$ and $\sigma$ across the evaluated inverse problems are detailed in \cref{tab:afhq-configs,tab:celeba-half-noise}. These configurations were determined empirically via a grid search on a held-out validation set.

\begin{table}[h]
\centering
\caption{Hyper-parameters for AFHQ Cat}
\label{tab:afhq-configs}
\begin{tabular}{lcccc}
\toprule
Problem & Steps & Init Mode & $\lambda$ & $\sigma$ \\
\midrule
Denoising & 20 & $A^\top$ & 0.04 & 0.08 \\
Gaussian Deblurring & 20 & $A^\top$ & 1.50 & 0.32 \\
Inpainting & 20 & Conv & 0.30 & 0.10 \\
Paintbrush Inpainting & 20 & Conv & 0.30 & 0.10 \\
Random Inpainting & 20 & Conv & 1.60 & 0.05 \\
Super-resolution & 20 & Bicubic & 2.20 & 0.15 \\
\bottomrule
\end{tabular}
\end{table}

\begin{table}[h]
\centering
\caption{Hyper-parameters for CelebA}
\label{tab:celeba-half-noise}
\begin{tabular}{lcccc}
\toprule
Problem & Steps & Init Mode & $\lambda$ & $\sigma$ \\
\midrule
Denoising & 20 & $A^\top$ & 0.10 & 0.05 \\
Gaussian Deblurring & 20 & $A^\top$ & 0.80 & 0.05 \\
Inpainting & 20 & Conv & 0.45 & 0.05 \\
Paintbrush Inpainting & 20 & Conv & 0.80 & 0.15 \\
Random Inpainting & 20 & Conv & 0.45 & 0.05 \\
Super-resolution & 20 & Bicubic & 1.20 & 0.05 \\
\bottomrule
\end{tabular}
\end{table}

Each experiment used a single NVIDIA L40S GPU or NVIDIA A40 GPU, each with 49GB of memory.

\subsection{Comparisons between different priors}

To ensure fair comparison between Spherical Encoder and diffusion/flow-based methods, we validate that the generative models used as the backbone for all methods mostly achieve similar KID results for generation, as illustrated in \cref{tab:prior_kid}.

\begin{table*}[h]
\centering
\begin{tabular}{lcc}
\toprule
Prior & AFHQ-Cat & CelebA \\
\midrule
Diffusion & 14.1 & 104.4$^*$\\
Flow & 10.78 & 11.4 \\
Sphere Encoder & 9.5 & 10.4 \\
\bottomrule
\end{tabular}
\caption{KID $\times 10^3$ of different priors on AFHQ-Cat and CelebA. $^*$Despite the high KID number achieved by the diffusion model prior~\cite{rdruce-2026-ddpm-celeb-128}, diffusion based methods (\textit{e.g.}, DPS) achieve state-of-the-art results.}
\label{tab:prior_kid}
\end{table*}

\subsection{Licences}

AFHQ~\cite{choi2020AFHQ} is released under CC BY-NC 4.0; CelebA~\cite{yang2015facial} is released under a custom non-commercial research license.

%% file: sections/A_proofs_v2.tex
\section{Theoretical Analysis and Convergence}
\label{app:theoretical analysis}

While classical Half-Quadratic Splitting guarantees convergence for specific regularizers \cite{doi:10.1137/030600862}, extending these guarantees to highly non-linear deep generative priors requires treating the alternating updates as a non-convex optimization problem. Under the assumption that the objective satisfies the Kurdyka-\L{}ojasiewicz property, alternating minimization schemes are guaranteed to converge to a critical point \cite{10.1007/s10107-013-0701-9}. More recently, similar theoretical extensions have been specifically proven for HQS-based Plug-and-Play architectures \cite{hurault2022gradient}.

To formally analyze the convergence behavior of \coolname~as a fixed-point iteration, we follow standard fixed-point and convex optimization theory~\cite{rockafellar2009variational, bauschke2020correction}. While the assumptions required for formal fixed-point convergence are restrictive and rarely strictly satisfied by deep generative models, they provide the mathematical framework that motivates our empirical analysis in \Cref{sec:method}.

\subsection{Preliminaries and Notations}

Let $\mathcal{X}=\mathbb{R}^n$ be a finite-dimensional image space, $\mathcal{V}=\mathbb{R}^m$ be a finite-dimensional latent space, and $\mathcal{Y}=\mathbb{R}^k$ be a finite-dimensional measurement space, with a linear forward operator $A:\mathcal{X} \to \mathcal{Y}$ containing a non-trivial null-space. 

Our algorithm, \coolname, updates the image state via the operator $T$:
\begin{equation}
 T(x)= \text{prox}_{\eta \ell}(P(x)), 
\end{equation}
where $P(x) = D(S(E(x)))$ is the generative prior mapping defined in the main text. The data-fidelity proximal operator is defined for a convex loss $\ell(x) = \lVert Ax-y \rVert^2$ as:
\begin{equation}
    \text{prox}_{\eta \ell}(z) = \underset{x\in \mathcal{X}}{\arg\min} \Big\{ \ell(x) + \frac{1}{2\eta} \lVert x-z \rVert^2 \Big\}.
\end{equation}
Note that setting $\eta = \frac{1}{2\lambda}$ exactly recovers the update step defined in \cref{eq:hqs_data}.

\textbf{Definition 1 (Nonexpansive and Averaged Operators).} An operator $\mathcal{T}: \mathcal{X} \to \mathcal{X}$ is nonexpansive if it is $1$-Lipschitz continuous ($\|\mathcal{T}(x) - \mathcal{T}(y)\| \le \|x - y\|$). It is $\alpha$-averaged if there exists a nonexpansive operator $R$ and a constant $\alpha \in (0, 1)$ such that $\mathcal{T} = (1-\alpha)I + \alpha R$. An operator is firmly nonexpansive if it is exactly $1/2$-averaged.

\textbf{Lemma 1 (Composition).} \textit{The composition of an $\alpha_1$-averaged operator and an $\alpha_2$-averaged operator is an $\alpha_3$-averaged operator, where $\alpha_3 \in (0, 1)$ \cite{bauschke2020correction}.}

\subsection{Proof of Convergence}

Because $\ell(x)$ is closed, proper, and convex, its proximal operator 
$\operatorname{prox}_{\eta \ell}$ is firmly nonexpansive ($1/2$-averaged). If we assume the deep prior $P$ acts as an averaged operator, we can guarantee convergence to a fixed point.

\textbf{Theorem 1 (Convergence to a fixed point).} \textit{Assume that the prior mapping $P$ is $\alpha$-averaged on $\mathcal{X}$ and the fixed-point set $\operatorname{Fix}(T) = \{x : T(x) = x\}$ is nonempty. Then, the sequence generated by $x_{k+1} = T(x_k)$ converges to a fixed point $x_\infty \in \operatorname{Fix}(T)$.}

\textit{Proof.} 
By Lemma 1, the composition of the $1/2$-averaged $\operatorname{prox}_{\eta \ell}$ and the $\alpha$-averaged $P$ means the update operator $T$ is $\beta$-averaged for some $\beta \in (0, 1)$. A fundamental property of a $\beta$-averaged operator is that for any $x, y \in \mathcal{X}$:
\begin{equation}
\|T(x) - T(y)\|^2 \le \|x - y\|^2 - \frac{1-\beta}{\beta} \|(I - T)(x) - (I - T)(y)\|^2.    
\end{equation}
Let $x^* \in \operatorname{Fix}(T)$. Substituting $y = x^*$ (noting $T(x^*) = x^*$) and our iteration $x_{k+1} = T(x_k)$, we obtain:
\begin{equation}\label{eq:fejer}
  \|x_{k+1} - x^*\|^2 \le \|x_k - x^*\|^2 - \frac{1-\beta}{\beta} \|x_k - x_{k+1}\|^2.  
\end{equation}
Because the subtracted term is non-negative, $\|x_{k+1} - x^*\| \le \|x_k - x^*\|$. Therefore, the sequence $\{x_k\}$ is Fejér monotone, bounded, and non-increasing in distance to any fixed point. 

Summing \Cref{eq:fejer} from $k=0$ to $K-1$ yields a telescoping sum bounding the residuals:
\begin{equation}
\frac{1-\beta}{\beta} \sum_{k=0}^{K-1} \|x_k - x_{k+1}\|^2 \le \|x_0 - x^*\|^2.
\end{equation}
Because the infinite sum is bounded by a finite constant, the individual terms must approach zero ($\|x_k - x_{k+1}\| \to 0$), demonstrating the sequence is asymptotically regular. In a finite-dimensional space, the Krasnosel'skii-Mann theorem dictates that a sequence which is both bounded and asymptotically regular under an averaged operator strictly converges to a point in the fixed-point set. Thus, $x_k \to x_\infty \in \operatorname{Fix}(T)$. $\blacksquare$

\textbf{Remark 1 (On Linear Convergence).} If the data fidelity term $\ell(x)$ was strongly convex, its proximal operator would be a strict contraction, yielding a linear convergence rate via the Banach Fixed Point Theorem. However, in typical image restoration tasks (\textit{e.g.}, deblurring, inpainting), the forward operator $A$ has a non-trivial null-space, meaning $\ell(x)$ is not strongly convex. Consequently, theoretical bounds on the convergence rate are difficult to establish, which motivates the empirical convergence validation provided in \cref{fig:residual}.